\documentclass[journal]{IEEEtran}

\usepackage{graphicx}
\usepackage{setspace}
\usepackage{epstopdf}

\usepackage{wrapfig}
\usepackage{caption}
\usepackage{subcaption}

\usepackage{diagbox}
\usepackage{color}

\makeatletter
\def\BState{\State\hskip-\ALG@thistlm}
\makeatother

\usepackage[english]{babel}
\usepackage[mathletters]{ucs}
\usepackage[utf8]{inputenc}
\usepackage[export]{adjustbox}
\usepackage[shortlabels]{enumitem} 

\usepackage{csquotes}
\usepackage{latexsym}
\usepackage{color} 
\usepackage{indentfirst}
\usepackage{graphicx}       
\usepackage{fancyhdr}
\usepackage{longtable}
\usepackage{pifont}
\usepackage{makeidx}
\usepackage{lastpage}
\usepackage{multirow}
\usepackage{dcolumn} 
\usepackage{epstopdf}
\usepackage{url}
\usepackage{listings}
\usepackage{relsize}
\usepackage{pdfpages}
\usepackage{tabularx}
\usepackage{booktabs}  
\usepackage{float}
\usepackage{textcomp}
\usepackage{hyperref}
\usepackage{afterpage}
\usepackage{bm}  
\usepackage[normalem]{ulem}

\usepackage[nolist,nohyperlinks]{acronym}
\usepackage{algorithm}
\usepackage{algpseudocode}

\usepackage{standalone}
\usepackage{pgfplots}
\usepackage{cuted}

\usepackage{diagbox}
\usepackage{amsmath,amssymb,amsthm}

\usepackage{nicefrac}

\usepackage{letltxmacro}

\usepackage{amssymb,amsmath}
\usepackage{mathtools}    
\usepackage[binary-units=true]{siunitx}
\usepackage{interval}

\newcommand{\expected}[1]{\mathrm{E}\left[ #1 \right]}
\newcommand{\variance}[1]{\mathrm{var}\left( #1 \right)}
\newcommand{\norm}[1]{\left\lVert#1\right\rVert}
\newcommand{\abs}[1]{\left\lvert#1\right\rvert}

\usepackage[e]{esvect}
\let\oldvec\vv
\renewcommand{\vec}[1]{\oldvec{\bm{#1}}}
\newcommand{\pnt}[1]{\bm{#1}}
\newcommand\mat{\mathbf}

\newcommand{\bemat}[1]
{
  \begin{bmatrix}
    #1
  \end{bmatrix}
}

\newcommand{\tstep}[2][]{_{#1[#2]}}
\newcommand*{\tran}{^{\intercal}}

\DeclareSIUnit{\pixel}{px}
\DeclareSIUnit{\fps}{FPS}

\DeclareMathOperator*{\argmin}{argmin}

\DeclareMathOperator{\diag}{diag}
\DeclareMathOperator{\trace}{tr}

\intervalconfig{soft open fences}

\pgfplotsset{compat=1.16}

\let\originalleft\left
\let\originalright\right
\renewcommand{\left}{\mathopen{}\mathclose\bgroup\originalleft}
\renewcommand{\right}{\aftergroup\egroup\originalright}

\newcommand{\prob}[2][]{p_{#1}\left(#2\right)}
\newcommand{\cprob}[3][]{\prob[#1]{#2 \mid #3}}

\newcommand{\set}[1]{\mathcal{\expandafter\MakeUppercase\expandafter{#1}}}

\newcommand{\vox}{v}
\newcommand{\mapval}{G}
\newcommand{\mapfun}{g}
\newcommand{\mapw}{n}
\newcommand{\map}{\set{M}}

\newcommand{\at}[2]{\left.\kern-\nulldelimiterspace#1\right|_{#2}}

\addto\extrasenglish{\def\equationautorefname~#1\null{eq.~(#1)\null}}
\newcommand{\rone}[1]{#1}
\newcommand{\rtwo}[1]{#1}

\usepackage{scalerel}
\usetikzlibrary{svg.path}
\definecolor{orcidlogocol}{HTML}{A6CE39}
\tikzset{
  orcidlogo/.pic={
    \fill[orcidlogocol] svg{M256,128c0,70.7-57.3,128-128,128C57.3,256,0,198.7,0,128C0,57.3,57.3,0,128,0C198.7,0,256,57.3,256,128z};
    \fill[white] svg{M86.3,186.2H70.9V79.1h15.4v48.4V186.2z}
    svg{M108.9,79.1h41.6c39.6,0,57,28.3,57,53.6c0,27.5-21.5,53.6-56.8,53.6h-41.8V79.1z M124.3,172.4h24.5c34.9,0,42.9-26.5,42.9-39.7c0-21.5-13.7-39.7-43.7-39.7h-23.7V172.4z}
    svg{M88.7,56.8c0,5.5-4.5,10.1-10.1,10.1c-5.6,0-10.1-4.6-10.1-10.1c0-5.6,4.5-10.1,10.1-10.1C84.2,46.7,88.7,51.3,88.7,56.8z};
  }
}

\newcommand\orcidicon[1]{\href{https://orcid.org/#1}{\mbox{\scalerel*{
        \begin{tikzpicture}[yscale=-1,transform shape]
          \pic{orcidlogo};
        \end{tikzpicture}
}{|}}}}

\acrodef{AAIS}[AAIS]{Autonomous Aerial Interception System}
\acrodefplural{AAIS}[AAIS']{Autonomous Aerial Interception Systems}

\acrodef{C-UAS}[C-UAS]{Counter UAV System}
\acrodefplural{C-UAS}[C-UAS']{Counter UAV Systems}

\acrodef{FoV}[FoV]{Field of View}
\acrodef{VFoV}[VFoV]{Vertical Field of View}
\acrodef{HFoV}[HFoV]{Horizontal Field of View}
\acrodef{ARL}[ARL]{Application Readiness Level}
\acrodef{BFS}[BFS]{Breadth-First Search}
\acrodef{UAV}[UAV]{Unmanned Aerial Vehicle}
\acrodef{GPS}[GPS]{Global Positioning System}
\acrodef{SLAM}[SLAM]{Simultaneous Localization And Mapping}
\acrodef{SLAMs}[SLAMs]{Simultaneous Localization And Mapping systems}
\acrodef{GPS}[GPS]{Global Positioning System}
\acrodef{RTK}[RTK]{Real-time Kinematic}
\acrodef{GNSS}[GNSS]{Global Navigation Satellite System}
\acrodef{ROS}[ROS]{Robot Operating System}
\acrodef{API}[API]{Application Programming Interface}
\acrodef{UGV}[UGV]{Unmanned Ground Vehicle}
\acrodef{UV}[UV]{Ultra-Violet}
\acrodef{LED}[LED]{Light-emitting Diode}
\acrodef{MBZIRC}[MBZIRC]{Mohamed Bin Zayed International Robotics Challenge}
\acrodef{DARPA}[DARPA]{Defense Advanced Research Projects Agency}
\acrodef{SAR}[SAR]{Search and Rescue}
\acrodef{IMU}[IMU]{Inertial Measurement Unit}
\acrodef{LTI}[LTI]{Linear time-invariant}
\acrodef{MPC}[MPC]{Model Predictive Control}
\acrodef{UVDAR}[UVDAR]{Ultra-Violet Direction And Ranging}
\acrodef{DOF}[DOF]{degree-of-freedom}
\acrodef{DOFs}[DOFs]{degrees-of-freedom}
\acrodef{LiDAR}[LiDAR]{Light Detection and Ranging}
\acrodef{ESC}[ESC]{Electronic Speed Controller}
\acrodef{ICP}[ICP]{Iterative Closest Points}
\acrodef{KF}[KF]{Kalman Filter}
\acrodef{LKF}[LKF]{Linear Kalman Filter}
\acrodef{UKF}[UKF]{Unscented Kalman Filter}
\acrodef{EKF}[EKF]{Extended Kalman Filter}
\acrodef{RAS}[RAS]{Robotics and Automation Society}
\acrodef{IEEE}[IEEE]{Institute of Electrical and Electronics Engineers}
\acrodef{MRS}[MRS]{Multi-robot Systems Group}
\acrodef{CNN}[CNN]{Convolutional Neural Network}
\acrodef{CTU}[CTU]{Czech Technical University}
\acrodef{UPenn}[UPenn]{University of Pennsylvania}
\acrodef{NYU}[NYU]{New York University}
\acrodef{FIFO}[FIFO]{First In, First Out}
\acrodef{RMSE}[RMSE]{Root Mean Square Error}
\acrodef{PDF}[PDF]{Probability Distribution Function}
\acrodef{CDF}[CDF]{Cumulative Distribution Function}
\acrodef{MC}[MC]{Monte-Carlo}
\acrodef{TSDF}[TSDF]{Truncated Signed Distance Field}
\acrodef{SWaP}[SWaP]{Size, Weight, and Power}
\acrodef{MLP}[MLP]{Multi-Layer Perceptron}

\LetLtxMacro{\oldinterval}{\interval}
\renewcommand{\interval}[2][]{\oldinterval[scaled, #1]{#2}}

\newcommand{\PREPRINTYEAR}{2025}
\newcommand{\PUBLISHEDIN}{IEEE Transactions on Robotics}
\newcommand{\DOI}{10.1109/TRO.2024.3502494} 

\usepackage[placement=top,vshift=-7]{background}
\SetBgScale{1.0}
\SetBgContents{\PUBLISHEDIN. PREPRINT VERSION - DO NOT DISTRIBUTE. \href{https://doi.org/\DOI}{DOI \DOI}}
\SetBgColor{black}
\SetBgAngle{0}
\SetBgOpacity{1.0}

\begin{document}

\thispagestyle{empty}
\onecolumn
{
  \topskip0pt
  \vspace*{\fill}
  \centering
  \LARGE{%
    \copyright{} \PREPRINTYEAR~\PUBLISHEDIN\\\vspace{1cm}
    Personal use of this material is permitted.
    Permission from \PUBLISHEDIN~must be obtained for all other uses, in any current or future media, including reprinting or republishing this material for advertising or promotional purposes, creating new collective works, for resale or redistribution to servers or lists, or reuse of any copyrighted component of this work in other works.}
    \vspace*{\fill}
}
\NoBgThispage
\twocolumn          	
\BgThispage


\title{On Onboard LiDAR-based Flying Object Detection}

\author{Matouš Vrba$^{*\dag}$$^{\orcidicon{0000-0002-4823-8291}}$, Viktor Walter$^{*}$$^{\orcidicon{0000-0001-8693-6261}}$, Václav Pritzl$^{*}$$^{\orcidicon{0000-0002-7248-6666}}$, Michal Pliska$^{*}$$^{\orcidicon{0009-0006-7205-0455}}$, Tomáš Báča$^{*}$$^{\orcidicon{0000-0001-9649-8277}}$, Vojtěch Spurný$^{*}$$^{\orcidicon{0000-0002-9019-1634}}$, Daniel~Heřt$^{*}$$^{\orcidicon{0000-0003-1637-6806}}$, and Martin Saska$^{*}$$^{\orcidicon{0000-0001-7106-3816}}$
  \thanks{%
      This work was supported by CTU in Prague grant no. SGS23/177/OHK3/3T/13, %
      by the Czech Science Foundation (GAČR) under research project no. 23-07517S, %
      and by the European Union under the project \enquote{Robotics and advanced industrial production} (reg. no. CZ.02.01.01/00/22\_008/0004590).%
  }%
  \thanks{$^{*}$Authors are with the Faculty of Electrical Engineering,
      Czech Technical University in Prague, Technická 2, Prague 6,
      (email: {\tt {\{matous.vrba, viktor.walter, vaclav.pritzl, michal.pliska, tomas.baca, vojtech.spurny, daniel.hert, martin.saska\}@fel.cvut.cz}}).}%
  \thanks{$^{\dag}$Corresponding author.}%
}

\maketitle

\begin{abstract}
A new robust and accurate approach for the detection and localization of flying objects with the purpose of highly dynamic aerial interception and agile multi-robot interaction is presented in this paper.
The approach is proposed for use on board of autonomous aerial vehicles equipped with a 3D \acs{LiDAR} sensor.
It relies on a novel 3D occupancy voxel mapping method for the target detection that provides high localization accuracy and robustness with respect to varying environments and appearance changes of the target.
In combination with a proposed cluster-based multi-target tracker, sporadic false positives are suppressed, state estimation of the target is provided, and the detection latency is negligible.
This makes the system suitable for tasks of agile multi-robot interaction, such as autonomous aerial interception or formation control where fast, precise, and robust relative localization of other robots is crucial.
\rone{We evaluate the viability and performance of the system in simulated and real-world experiments which demonstrate that at a range of \SI{20}{\metre}, our system is capable of reliably detecting a micro-scale \acs{UAV} with an almost \SI{100}{\percent} recall, \SI{0.2}{\metre} accuracy, and \SI{20}{\milli\second} delay.}
\end{abstract}
\begin{IEEEkeywords}
  	Aerial Systems: Perception and Autonomy; Multi-Robot Systems; Object Detection, Segmentation and Categorization; Autonomous Aerial Interception
\end{IEEEkeywords}



\section{Introduction}

\begin{figure}
  \centering
  \includegraphics[width=0.48\textwidth]{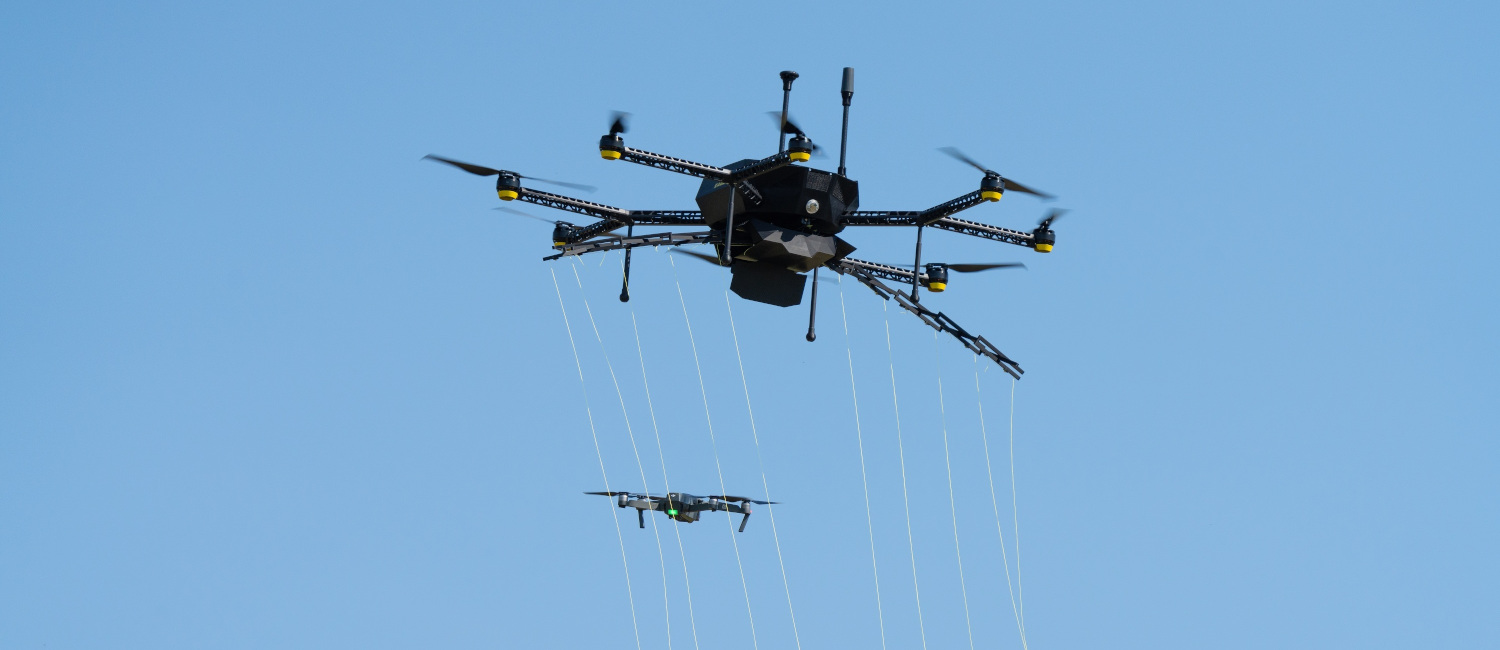}
  \caption{%
    The proposed detection method deployed on board an autonomous aerial interception system while eliminating an intruding DJI Mavic \ac{UAV}.
  }
  \label{fig:eagle_catch}
\end{figure}

With the recent rise in the popularity, availability, and utility of multirotor and fixed-wing \acp{UAV}, there exists a growing concern regarding aerial safety.
Reviews of recent \ac{UAV}-related malpractices and accidents, such as the famous Gatwick Airport incident of 2018 when the airport was closed for three days due to a reported \ac{UAV} sighting, are provided in \cite{chamola2021attacks}, \cite{chiper2022survey} and \cite{ghasri2021accidents}, which all conclude that the current aerial safety measures are insufficient to deal with unmanned vehicles.
In this work, we focus on autonomous aerial interception, which provides several key advantages over a manual or ground-based \ac{C-UAS}.
Most ground-based \ac{C-UAS} rely on spoofing or jamming of the intruding \ac{UAV}'s navigation or radio-control signals, or on physical takedown of the \ac{UAV}, which can potentially cause an uncontrolled landing or a crash of the target, endangering the people or equipment below the intruder.
An \ac{AAIS} can safely capture and dispose of an intruding \ac{UAV} using non-destructive means (eg. an onboard net as illustrated in Fig.~\ref{fig:eagle_catch}), and the intruder can then also be used for forensics and investigation.
Furthermore, because such systems can work fully or semi-autonomously, they are not prone to human error or limited by human reflexes and do not require an unobstructed line of sight between the human operator and the flying target.

However, as reported in~\cite{stampa2021maturity}, the maturity of fully autonomous \ac{C-UAS}s is lacking.
The authors rate the maturity of manual \ac{C-UAS} systems with an \ac{ARL} of 7 out of 9, which corresponds to \enquote{functionality demonstrated}, but not proven.
The maturity of autonomous \ac{C-UAS} systems was rated only with an \ac{ARL} of 6, corresponding to \enquote{potential demonstrated}.
Based on our experience with \ac{AAIS} research, we conclude that this is mostly due to the lack of a suitable system for detection and localization of the target \ac{UAV} \cite{vrba_ral2019}, \cite{vrba_ral2020}.
Such a system has to be accurate, robust to false positives, various target appearances, and different environments, and provide high-frequency  updates with a low delay in order to enable elimination of agile maneuvering targets.
To fulfill these conditions and also to ensure the security and reliability of the system, the detector should run on board the interceptor \ac{UAV} without relying on external sensors and communication.
A large portion of the recent research tackling this problem was inspired by the MBZIRC 2020 international robotic competition\footnote{\href{http://mrs.felk.cvut.cz/mbzirc2020}{\url{http://mrs.felk.cvut.cz/mbzirc2020}}} where one of the challenges emulated an aerial autonomous interception scenario~\cite{mbzirc2020ch1_ollero, mbzirc2020ch1_behnke, mbzirc2020ch1_inaba, barisic2022mbzirc}.
Complexity of the problem is well illustrated by the fact that only four teams of the 22 expert competitors were able to intercept the target, even under the simplified conditions in the controlled environment of the competition.
To our best knowledge, no complete and reliable solution to \ac{UAV} detection and localization for autonomous aerial interception has been presented yet, and more research and development is still required.

Inspired by our solution to the MBZIRC 2020 competition, which was one of the four successful approaches and scored second place \cite{vrba_ras2022}, we propose a new method to address the detection problem.
The proposed method relies on a \ac{LiDAR} sensor placed on board the interceptor \ac{UAV} and has already been employed in implementation of the prototype \ac{AAIS} platform Eagle.One\footnote{\href{https://eagle.one}{\url{https://eagle.one}}, \href{https://mrs.felk.cvut.cz/projects/eagle-one}{\url{https://mrs.felk.cvut.cz/projects/eagle-one}}} (pictured in Fig.~\ref{fig:eagle_catch}), which is capable of autonomously eliminating a flying target~\cite{pliska2024towards}.
It is also suitable for general multi-robot applications where relative localization of the robots is required as evidenced by its deployment in a cooperative \ac{UAV} navigation scenario in our previous work \cite{pritzl2022icuas}.

\subsection{Related works}
\label{sec:sota}
Surveys of state-of-the-art \ac{C-UAS} systems~\cite{chamola2021attacks}, \cite{chiper2022survey} rarely consider solutions based on \ac{LiDAR} detection, which are relatively sparse when compared to other detection methods, such as RADAR~\cite{park2019radar, quevedo2018radar}, visual~\cite{vrba_ral2020}, acoustic~\cite{shi2020acoustic} or even using mixed sensors~\cite{svanstrom2020mixed}.
Most of the existing solutions for detecting \acp{UAV} are ground-based and rely on the assumption that the sensor is stationary, which also applies to \ac{LiDAR}-based methods, such as~\cite{hammer2019lidar_detection} or~\cite{dogru2022lidar_detection}.
However, these approaches are not suitable for an autonomous interception, where low delay and high accuracy localization of the target must be provided, potentially over a large area.
The requirement for the detector to run on board a \ac{UAV} disqualifies methods that rely on a large or heavy sensor, such as most of the RADAR-based approaches, as well as methods assuming a static sensor or low acoustic background noise.

Visual cameras can be small and lightweight and are generally suitable for deployment on board \acp{UAV}.
There are plenty of works on \ac{UAV} detection from RGB images using \acp{CNN} or conventional computer vision methods, such as \cite{vrba_ral2020}, \cite{barisic2022sim2air}, \cite{shumann2017}, and \cite{rozantsev2017detection}.
These are useful for long-range detection of the target, but do not provide sufficiently accurate localization necessary for its physical elimination.
For accurate detection and localization at close ranges, depth information from a stereo camera can be used either to complement an RGB detector for the localization as in \cite{barisic2022mbzirc}, or to directly detect and localize the target as in~\cite{vrba_ral2019}, \cite{carrio2020}.
However, these approaches only consider a single depth or RGB image at once and do not take full advantage of the spatial information provided by the sensor over time, leading to false positives when the current scene is ambivalent.

This also applies to \ac{LiDAR}-based methods for \ac{UAV} detection.
Ground-based detectors, such as~\cite{dogru2022lidar_detection} typically leverage the assumption of a non-moving sensor and a static background to remove points corresponding to the background, which enables detecting the target by Euclidean clustering of the remaining points.
However, such background removal is not applicable to deployment onboard a moving \ac{UAV} in general environments.
On the other hand, onboard methods mostly assume an obstacle-free environment, such as a sufficiently high flight altitude, which leads to similar detection algorithms as the ground-based static solutions after background removal~\cite{vrba_ras2022}, \cite{aldao2022lidar_detection_onboard}.
These approaches are unsuitable for more complex applications, such as \ac{C-UAS} using \ac{AAIS} or aerial multi-robot cooperation, where such conditions cannot be assumed.

\rone{%
During the last years, deep learning-based techniques for feature extraction and detection in point clouds were developed, following the successes of \acp{CNN} for computer vision tasks.
Early works attempted to directly transfer \acp{CNN} to 3D using generalization of the 2D convolution to 3D voxels or by projecting the 3D data to a 2D image~\cite{qi2016VolumetricMultiViewCNNs}.
In~\cite{qi2017PointNetDeepLearning}, the authors introduced a new approach for neural network-based point cloud processing dubbed the PointNet, which is now widely used as a building block of other architectures.
The PointNet++ builds upon this approach and introduces a hierarchical feature fusion similar to \acp{CNN} to obtain features at different levels of abstraction~\cite{qi2017PointNet2}.
PointNet-based feature extraction was combined with the Hough transform in~\cite{qi2019DeepHoughVoting} to design the VoteNet 3D object detector.
The Hough voting was shown to improve the detection performance over a conventional \ac{MLP} using the same feature extractor.
A more recent work proposes the Point-GNN~\cite{shi2020PointGNNGraphNeural} detector, which relies on a graph-based representation of the input point cloud.
The authors show that this is a viable alternative to the \ac{CNN}-based and PointNet-based deep learning approaches.
Although the neural networks show impressive results in general classification, detection, or semantic segmentation tasks, these do not easily transfer to the problem addressed in this work as we demonstrate in sec.~\ref{sec:deep-learning}.
}

Overall, the state-of-the-art detection methods are either not suitable for deployment on board aerial robots, not sufficiently fast, accurate, and robust, or cannot be employed in general environments with obstacles.
\rone{To address the issues of robustness and generality, we propose to integrate the spatial information sampled by the onboard sensors using a dynamic occupancy map of the environment that is then used to detect the flying objects.
Including information from previous spatial observations helps to remove ambiguities in the observed scene, but it also makes the detection algorithm computationally demanding, limiting the detection rate.
To tackle this, we design a specialized, computationally lightweight tracking algorithm.
The tracker is initialized and updated using low-rate detections and then utilizes the latest sensory data directly, which ensures a low delay and high rate of the tracking output.
The resulting pipeline overcomes the limitations of the state of the art stated above.
To the best of our knowledge, few object detectors integrate spatial information over time, and none can easily be applied to the problem of onboard marker-less flying object detection.
Furthermore, we also provide a deep theoretical analysis of relative position measurements on board a platform with pose uncertainty, which we argue is an important yet often overlooked aspect of mapping and relative localization in mobile robotics.}

\begin{figure*}
  \centering
  \includegraphics[width=\textwidth]{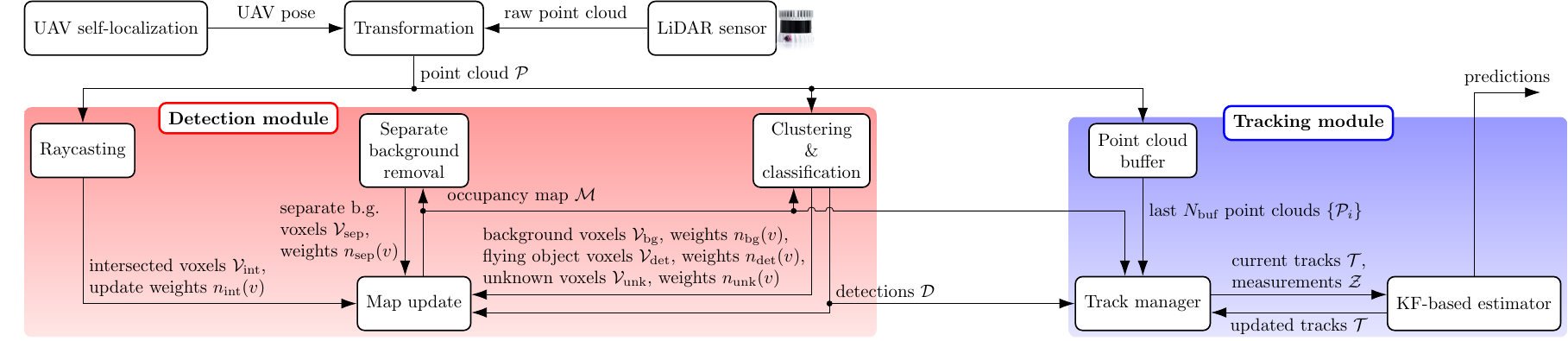}
  \caption{%
    Schematic overview of the detection and tracking method proposed in this paper.
  }
  \label{fig:schematic_overview}
\end{figure*}

In the context of integrating spatial information, our work is related to occupancy mapping algorithms.
Representative examples are the widely used OctoMap~\cite{hornung2013octomap} and the newer VoxBlox~\cite{oleynikova2017voxblox} mapping libraries.
Both represent the environment using voxels stored in an efficient manner, whose values store information about occupancy of the respective voxel in case of the OctoMap, or about the nearest obstacle in case of the VoxBlox.
These values are updated by integrating the incoming point clouds and raycasting the corresponding \ac{LiDAR} rays intersecting free space.
An alternative to voxel-based mapping is direct integration of the measured point clouds over time, which can provide fast collision checking for trajectory planning, such as the approach presented in~\cite{kong2021avoiding}. 
However, neither of these approaches explicitly considers dynamic obstacles and none are able to distinguish flying objects from the background, so a novel mapping method had to be designed.

We summarize our contributions as follows:
\begin{enumerate}
  \item a novel occupancy mapping approach that explicitly encodes unknown space, reacts quickly to dynamic obstacles and supports detection of flying objects,
  \item an accurate and robust flying object detector for \acp{UAV} that takes advantage of the current and past spatial information sampled by the onboard sensor,
  \item a multi-target tracking module to reduce the detection delay and to provide association, state estimation, and prediction of the targets,
  \item a theoretical analysis of properties and limitations of the proposed methods with applications to general mapping and object localization,
  \item a new dataset for \ac{UAV} detection in point clouds,
  \item and an extensive experimental evaluation and demonstration in both simulated and real-world scenarios including autonomous aerial interception.
\end{enumerate}
Additionally, the whole system is fully open-source and available online\footnote{MRS UAV System: \href{https://github.com/ctu-mrs/mrs_uav_system}{\url{https://github.com/ctu-mrs/mrs_uav_system}}}
\footnote{Detector: \href{https://github.com/ctu-mrs/vofod}{\url{https://github.com/ctu-mrs/vofod}}}
\footnote{Multi-target Tracker: \href{https://github.com/ctu-mrs/lidar_tracker}{\url{https://github.com/ctu-mrs/lidar_tracker}}}
 including the dataset\footnote{Dataset: \href{https://mrs.felk.cvut.cz/flying-object-detection}{\url{https://mrs.felk.cvut.cz/flying-object-detection}}}.



\section{Notation}
In this text, $\map$ is used to denote a voxel grid map, $\vox$ denotes a voxel in $\map$, $\set{V}$ denotes a set of voxels, $\set{P}$ denotes a set of 3D points from a single scan of the LiDAR sensor (i.e. a point cloud), $\set{C}$ denotes a cluster (subset) of points from $\set{P}$, and $\pnt{p}$ denotes a point from $\set{P}$ or $\set{C}$.
Centroid of a cluster $\set{C}$ (defined as the mean of $\pnt{p} \in \set{C}$) is denoted $\pnt{c}\left( \set{C} \right)$, and the number of elements in a set $\set{S}$ is denoted $\abs{\set{S}}$.

Furthermore, we define these relations between points and voxels as follows: $\vox(\pnt{p})$ is the voxel containing the point $\pnt{p}$, $\set{P} \cap \vox$ is a subset of points from $\set{P}$ contained within $\vox$, and $\pnt{c}_\vox$ is the center point of $\vox$.
When referring to a distance between two voxels $\vox_1$ and $\vox_2$ or a voxel-point distance between $\vox$ and $\pnt{p}$, the Euclidean norm $\norm{\pnt{c}_{\vox_1} - \pnt{c}_{\vox_2}}$ or $\norm{\pnt{c}_{\vox} - \pnt{p}}$ is intended, respectively.

The value of a variable at a certain time-step $t$ is indicated with a subscript~${}\tstep{t}$ of the variable.
For brevity, time-step subscripts are omitted unless they are relevant.



\section{Flying objects detection algorithm}
\label{sec:method}

An overview of the detection system is presented in Fig.~\ref{fig:schematic_overview}.
The main element of the system is a \textbf{Detection module} that relies on an occupancy voxelmap $\map$ with voxel size $s_v$.
Scans from the LiDAR are transformed to a static world frame based on the \ac{UAV}'s self-localization pipeline.
The transformed scans $\set{P}$ are used to iteratively update $\map$ as follows.

Firstly, points in $\set{P}$ are separated into Euclidean clusters and classified as one of three categories based on the surrounding voxels from $\map$: \textit{background}, \textit{flying object}, and \textit{unknown}.
Voxels containing these points are updated using weights that are obtained as the number of points within each voxel.
Secondly, a raycasting algorithm determines voxels intersected by the LiDAR's rays and their update weights as the total length of rays within each voxel.
Voxels in the map are updated either as \textit{occupied} or \textit{unknown} depending on the class of the corresponding cluster, or as \textit{free} for the voxels intersected by a ray.

Using this approach, each voxel converges to one of four states: \textit{confident occupied}, \textit{tentative occupied}, \textit{uncertain}, and \textit{confident free}.
To account for dynamic objects and to prevent \enquote{trails} of occupied voxels left behind such objects, a separate background removal algorithm periodically processes the map.
The algorithm detects voxel clusters that are classified as \textit{tentative occupied}, but are not connected to any \textit{confident occupied} voxels and updates these as \textit{uncertain} to remove such trails.
Finally, centroids of the point clusters classified as \textit{flying objects} are the output detections $\set{D}$ of the \textbf{Detection module}.

A \textbf{Tracking module} is used to associate subsequent detections corresponding to the same objects, to estimate and predict the targets' states, and to compensate the processing delay of the detector.
It receives the point clouds $\set{P}$, detections $\set{D}$, and the latest occupancy map $\map$.
A buffer keeps the last $N_{\text{buf}}$ point clouds as sorted by the time of acquisition.
When a new set of detections $\set{D}$ is obtained, the corresponding point cloud is selected in the buffer ($\set{D}$ may be delayed by several iterations of the sensor's update rate).
Each detection is then tracked through subsequent point clouds in the buffer to the latest one using a \ac{KF}-based multi-target tracking algorithm.
The multi-target tracker also uses the latest map $\map$ to prevent track association to background points.

All of these algorithms are implemented using the PCL~\cite{pcl} and Eigen~\cite{eigen} libraries in C++ and within the ROS framework~\cite{ros} and the MRS UAV system~\cite{baca2021mrs}.
They run in parallel to leverage the multi-core architectures of modern CPUs.
As mentioned in sec.~\ref{sec:sota}, the source code is available online.
Detailed descriptions of the algorithms follow below.















\subsection{Environment occupancy mapping and representation}
Traditional mapping algorithms typically use the log-odds representation of occupancy, as introduced in~\cite{moravec85logodds}. Therein, each cell $\vox$ in the map at time-step $t$ stores a single value
\begin{equation}
  L(\vox \mid z\tstep{1:t}) = \log\left( \frac{ \cprob[\text{occ}]{\vox}{z\tstep{1:t}} }{ 1 - \cprob[\text{occ}]{\vox}{z\tstep{1:t}}} \right), \label{eq:L}
\end{equation}
where $\cprob[\text{occ}]{\vox}{z\tstep{1:t}}$ is an estimate of the probability that cell $\vox$ is occupied based on the measurements $z\tstep{1:t}$.
Values of the cells are updated with each new point cloud $\mathcal{P}$ using points $\pnt{p} \in \mathcal{P}$ and the \ac{LiDAR}'s rays.
The update rule is based on the Markov assumption of history-independence and the Bayes' theorem for posterior probability,
\begin{equation}
  \cprob{A}{B} = \frac{ \cprob{B}{A} \prob{A} }{ \prob{B} }. \label{eq:bayes}
\end{equation}
Expanding \autoref{eq:bayes} for $\cprob[\text{occ}]{\vox}{z\tstep{1:t}}$ results in a complex expression containing several variables that are difficult to quantify in practice.
Using the probability-odds representation cancels these variables out and simplifies the update rule.
Assuming a prior occupancy probability $\prob[0,\text{occ}]{\vox} = 0.5$ and using a logarithm of the odds for numerical stability as in \autoref{eq:L}, the update rule
\begin{align}
  L\left( \vox \mid z\tstep{1:t} \right) = L\left( \vox \mid z\tstep{1:t-1} \right) + L\left( \vox \mid z\tstep{t} \right)
\end{align}
is obtained.
This update rule is typically parametrized by two values, $l_{\text{occ}}$ and $l_{\text{free}}$, as
\begin{align}
    L( \vox \mid z\tstep{t} ) =
    \begin{cases}
      l_{\text{occ}} & \text{if } \vox \text{ contains any point $\pnt{p} \in \mathcal{P}$}, \\
      l_{\text{free}} & \text{else if a ray passes through } \vox, \\
      0 & \text{otherwise}.
    \end{cases}
\end{align}
The state of a voxel $\vox$ can be discretely classified (e.g. for planning or compression purposes) based on predefined thresholds $L_{\text{occ}}$, $L_{\text{free}}$ as
\begin{align}
  \text{state}( \vox ) =
    \begin{cases}
      \text{confident occupied}   & \text{if } L(\vox) \geq L_{\text{occ}}, \\
      \text{confident free}       & \text{if } L(\vox) \leq L_{\text{free}}, \\
      \text{uncertain}            & \text{otherwise},
    \end{cases}
\end{align}
where $L(\vox)$ is short for $L(\vox \mid z\tstep{1:t})$.
To prevent windup and improve reaction time to dynamic obstacles, the value of $L(\vox)$ is often saturated to a minimal and maximal value, $L_{\text{min}}$ and $L_{\text{max}}$, respectively.
This limits the maximum number of updates necessary to change the voxel's state.

This model is commonly employed in many mapping algorithms~\cite{hornung2013octomap}, \cite{duberg2020ufomap}, \cite{meadhra2019varres} and has proved to be effective for mapping static environments, as demonstrated by the many applications relying on it~\cite{petracek2021caves}, \cite{kratky2021exploration}, \cite{xie2020map}.
However, it does not consider dynamic objects in the environment, which is crucial for their detection.
Furthermore, this model does not allow to easily increase the uncertainty of a cell, because repeated integration of the same type of measurement will always result in the cell's log-odds to eventually converge either towards a \textit{confident occupied} or a \textit{confident free} value.
As will be shown later in this paper, this makes the model unsuitable for the detection of flying objects.

\rone{%
To address these limitations, a different voxel occupancy representation and update rule are proposed in this paper.
The proposed representation allows for a more nuanced classification of the voxels in comparison to conventional mapping of static environments.
Our update rule takes into account the type of points used for updating the voxels, as classified by the method described in sec.~\ref{sec:clustering}.
To distinguish the two models, the value of a cell $\vox$ at time-step $t$ is denoted $\mapval\left( \vox \mid z\tstep{1:t} \right)$, or shortly $\mapval\left( \vox \right)$, when using the proposed representation.
Note that the interpretation of $L(\vox)$ and $G(\vox)$ is different and cannot be directly compared, but the interpretation of the corresponding state is analogous.

Similarly as the thresholds $L_{\text{occ}}$, $L_{\text{free}}$, for the log-odds representation, we define thresholds of a voxel's value $\mapval(\vox)$ corresponding to discrete classes of the voxel $v$.
The state classification is then
\begin{align}
  \text{state}( \vox ) =
    \begin{cases}
      \text{confident occupied} & \text{if } \mapval(\vox) \geq \mapval_{\text{conf}}, \\
      \text{tentative occupied} & \text{if } \mapval(\vox) \in \interval[open right]{\mapval_{\text{tent}}}{\mapval_{\text{conf}}}, \\
      \text{uncertain}          & \text{if } \mapval(\vox) \in \interval[open right]{\mapval_{\text{unc}}}{\mapval_{\text{tent}}}, \\
      \text{confident free}     & \text{if } \mapval(\vox) < \mapval_{\text{unc}}.
    \end{cases}
\end{align}
To update $\mapval(\vox)$ based on a new measurement $z\tstep{t}$, an exponential filter is employed in the form
\begin{equation}
  \mapval\left( \vox \mid z\tstep{1:t} \right) = \frac{ \mapval\left( \vox \mid z\tstep{1:t-1} \right) + \mapval\left( \vox \mid z\tstep{t} \right) }{ 2 }, \label{eq:update1}
\end{equation}
where $\mapval\left( \vox \mid z\tstep{t} \right)$ is the update value corresponding to the measurement.
Analogously to the parametrization $l_{\text{occ}}, l_{\text{free}}$ of the log-odds update, we define a set of constant update coefficients from which one is selected based on the measurement $z\tstep{t}$ and the voxel $\vox$.
For the flying object detection, we use
\begin{align}
    \mapval( \vox \mid z\tstep{t} ) =
    \begin{cases}
      \mapfun_{\text{occ}} & \text{ if } \vox \in \set{\vox}_{\text{bg}}, \\
      \mapfun_{\text{unk}} & \text{ if } \vox \in \set{\vox}_{\text{unk}}  \text{ or } \vox \in \set{\vox}_{\text{det}}, \\
      \mapfun_{\text{free}} & \text{ if } \vox \in \set{\vox}_{\text{int}} \text{ or } \vox \in \set{\vox}_{\text{sep}},
    \end{cases}
\end{align}
where $\set{\vox}_{\text{bg}}$, $\set{\vox}_{\text{unk}}$, $\set{\vox}_{\text{det}}$, $\set{\vox}_{\text{int}}$, and $\set{\vox}_{\text{sep}}$ are outputs of the respective submodules (see Fig.~\ref{fig:schematic_overview}) detailed below.
}%
If $\vox$ contains no point and no ray, $\mapval(\vox)$ remains unchanged.
It is assumed that a single voxel can only contain points of a single class (refer to sec.~\ref{sec:clustering}).
Note, that voxels containing points corresponding to a detected flying object are updated using $\mapfun_{\text{unk}}$ and not $\mapfun_{\text{free}}$.
This is to prevent biasing the map towards free voxels in the case of false-positive detections.

If $\vox$ contains multiple points, the update is applied per each point within the specific voxel.
\rone{Applying \autoref{eq:update1} $n$ times using the same update coefficient $\mapfun \in \left\{ \mapfun_{\text{occ}}, \mapfun_{\text{unk}}, \mapfun_{\text{free}} \right\}$ to update a voxel $\vox$ results in
\begin{equation}
  \mapval\left( \vox \mid z\tstep{1:t} \right) =
    \frac{
      \begin{aligned}
      \frac{
        \mapval\left( \vox \mid z\tstep{1:t-1} \right)
      }{2} + \frac{ \mapfun }{ 2 } \hspace{3.5em}\\[-0.9em]
        \vdots \hspace{3.35em} \hspace{2.1em} {}^{\ddots} + \frac{ \mapfun }{ 2 }
      \end{aligned}
    }{2} + \frac{ \mapfun }{ 2 },
\end{equation}
which can be rewritten as
\begin{equation}
  \mapval\left( \vox \mid z\tstep{1:t} \right) = \frac{\mapval\left( \vox \mid z\tstep{1:t-1} \right)}{ 2^n } + \frac{\mapfun}{2^n} + \frac{\mapfun}{2^{n-1}} + \cdots + \frac{\mapfun}{2}.
\end{equation}
The fractions of $\mapfun$ can be summed as a geometric progression to obtain
\begin{equation}
  \mapval\left( \vox \mid z\tstep{1:t} \right) = 2^{-n}\mapval\left( \vox \mid z\tstep{1:t-1} \right) + \left( 1 - 2^{-n} \right)\mapfun. \label{eq:update}
\end{equation}}
The \autoref{eq:update} allows for updating $\vox$ containing multiple points in a single calculation instead of iteratively.
More generally, instead of an integer number of updates, we use $\mapw \in \mathbb{R}$. This can be interpreted as a weighting coefficient associated with the update constant $\mapfun \in \left\{ \mapfun_{\text{occ}}, \mapfun_{\text{unk}}, \mapfun_{\text{free}} \right\}$.
The calculation of $\mapw$ and $\mapfun$ for a voxel $\vox$ during one iteration of the detection algorithm is described in sections~\ref{sec:clustering} and~\ref{sec:raycasting}.

\begin{figure}
  \centering
  \includegraphics[width=0.48\textwidth]{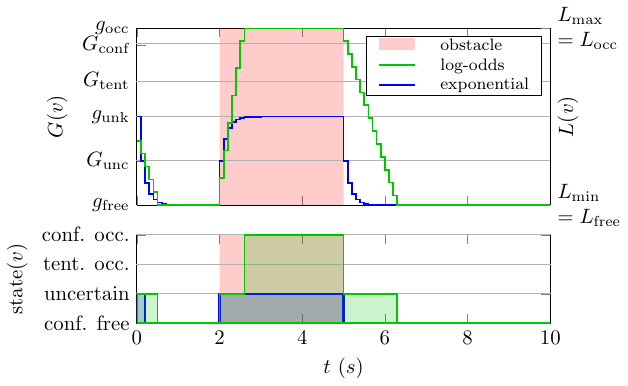}
  \caption{%
    Comparison of $L(\vox)$ and $\mapval(\vox)$ occupancy representations in a situation with a flying object entering an otherwise empty voxel at time $t=\SI{2}{\second}$ and leaving at $t=\SI{4.5}{\second}$.
    Values of the update and threshold parameters are chosen in the same way as suggested in~\cite{hornung2013octomap} for $L(\vox)$, and the same as in the real-world experiments for $\mapval(\vox)$.
    The current state according to both representations is presented in the lower graph.
  }
  \label{fig:graph_update_rule}
\end{figure}

The conventional log-odds voxel occupancy representation $L(\vox)$ and our multi-class exponential-update representation $\mapval(\vox)$ are compared in a simulated scenario where a dynamic flying obstacle enters a voxel.
Assuming that points corresponding to the obstacle are correctly identified as a \textbf{flying object}, our method never misclassifies the voxel as \textit{confidently occupied} (unlike the log-odds representation) and reacts faster to the object leaving the voxel (see Fig.~\ref{fig:graph_update_rule}).
These properties are crucial for the functioning of the detector presented further in this paper, but they can also benefit standard occupancy mapping applications with dynamic obstacles.


\subsection{Clustering \& classification}
\label{sec:clustering}
\begin{figure}
  \centering
  \includegraphics[width=0.48\textwidth]{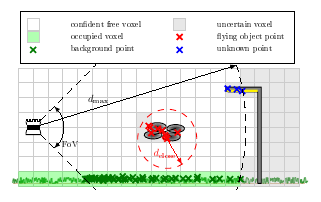}
  \caption{%
    Illustration of the classification algorithm on three clusters (red, green, and blue) of points (marked with crosses) within the sensor's \ac{FoV} and range (denoted $d_{\text{max}}$).
    Some points from the green cluster are closer than $d_{\text{close}}$ to \textit{occupied} voxels, so the cluster is classified as \textbf{background}.
    All points from the red cluster are further than $d_{\text{close}}$ from any \textit{occupied} voxels and are surrounded by \textit{confident free} voxels, so the cluster is classified as a \textbf{flying object}.
    Finally, the blue points are not close to \textit{occupied} voxels, but they are not decidedly separated from \textit{occupied} voxels by free air due to the unobserved part of the environment (which contains a connection of the object to the ground in this case).
    Therefore, it cannot be determined whether the blue points correspond to the background or a flying object at this point, so the cluster is classified as \textbf{unknown}.
  }
  \label{fig:cluster_classification}
\end{figure}

The algorithm outputs the set of occupied voxels $\mathcal{V}_{\text{occ}}$, their weighting coefficients $\mapw_{\text{occ}}(\vox)$, and a set of detections $\set{D}$ for each new point cloud $\mathcal{P}$.
Points from $\mathcal{P}$ are separated into clusters based on their mutual Euclidean distance using the method from~\cite{rusu_thesis}, such that for any two clusters $\mathcal{C}_k,\,\mathcal{C}_l,\,k \neq l$, these conditions hold:
\begin{align}
  \mathcal{C}_k \cap \mathcal{C}_l &= \emptyset,\\
  \forall \left\{ \pnt{p}_i \in \mathcal{C}_k,~\pnt{p}_j \in \mathcal{C}_l \right\},~\norm{\pnt{p}_i -  \pnt{p}_j} &> d_{\text{cluster}},
\end{align}
and for any two points $\pnt{p}_i,\,\pnt{p}_j \in \mathcal{C}_k$, there exists a subset $\left\{ \pnt{p}_{c}, \pnt{p}_{c+1}, \dots, \pnt{p}_{c+n} \right\} \subset \mathcal{C}_k$ such that
\begin{align}
  \begin{split}
    \norm{\pnt{p}_{i} -  \pnt{p}_{c}} &\leq d_{\text{cluster}},\\
    \norm{\pnt{p}_{c} -  \pnt{p}_{c+1}} &\leq d_{\text{cluster}},\\
    \dots,\\
    \norm{\pnt{p}_{c+n} -  \pnt{p}_{j}} &\leq d_{\text{cluster}}.\\
  \end{split}
\end{align}
Each cluster $\mathcal{C}$ is then classified using the following rules (illustrated in Fig.~\ref{fig:cluster_classification}):
\begin{enumerate}[label=\textbf{\Alph*)}]
  \item
    If at least one point from the cluster is closer than a threshold distance $d_{\text{close}}$ to any voxel that is at least \textit{tentative occupied}, the cluster is classified as \textbf{background}.\label{case:clusters_bg}
  \item If \ref{case:clusters_bg} is not met and the cluster is totally separated by \textit{confident free} voxels from any \textit{tentative} or \textit{confident occupied} voxels, it is classified as a \textbf{flying object}.
    To limit computational time, this condition is only checked within a sphere with radius $d_{\text{search}}$.
    \label{case:clusters_fod}
  \item If \ref{case:clusters_bg} does not hold and \ref{case:clusters_fod} neither or \ref{case:clusters_fod} cannot be determined within the $d_{\text{search}}$ radius, the cluster is classified as \textbf{unknown}.\label{case:clusters_unk}
\end{enumerate}

For a single point $\pnt{p}$, the condition \ref{case:clusters_bg} is evaluated by iterating through all of the voxels within a cube centered on $\vox(\pnt{p})$ with edge length $l = 2 d_{\text{close}} + 1$, checking if $\norm{ \pnt{c}_{\vox} - \pnt{p} } < d_{\text{close}} $ and $\mapval(\vox) \geq \mapval_{\text{conf}}$ holds for any of them.
This is repeated for each point and can be trivially implemented in the voxel grid map representation with a computational complexity $O(\abs{\set{C}} l^3)$.

The condition \ref{case:clusters_fod} is evaluated using Algorithm~\ref{alg:floodfill}.
The algorithm is a \ac{BFS} flood-fill approach that searches for the shortest path from a starting voxel to any \textit{tentative occupied} voxel that is not obstructed by \textit{confident free} voxels.
If such a path is found for any $v(\pnt{p}),~\pnt{p} \in \mathcal{C}$, the algorithm terminates and \ref{case:clusters_fod} is not met for $\mathcal{C}$ (lines \ref{alg:floodfill:if1start}-\ref{alg:floodfill:if1end} of the algorithm).
To prevent slowing down the algorithm when $\map$ contains many \textit{uncertain} voxels, the exploration is also terminated after reaching a border of a maximal-search sphere determined by its edge length $d_{\text{search}}$ (lines \ref{alg:floodfill:if2start}-\ref{alg:floodfill:if2end}), where $d_{\text{search}}$ is empirically determined based on the expected maximal dimensions of the targets.
If the algorithm fails to find a path to either a \textit{tentative occupied} voxel or the sphere's border for any $v(\pnt{p}),~\pnt{p} \in \mathcal{C}$, all points from the cluster are enclosed by \textit{confident free} voxels and \ref{case:clusters_fod} is satisfied.


{
\begin{algorithm}
  \algdef{SE}[SUBALG]{Indent}{EndIndent}{}{\algorithmicend\ }%
  \algtext*{Indent}
  \algtext*{EndIndent}

  \algnewcommand\AND{\textbf{and}~}
  \algnewcommand\Not{\textbf{not}~}
  \algnewcommand\Or{\textbf{or}~}
  \algnewcommand\algorithmicinput{\textbf{Input:~}}
  \algnewcommand\algorithmicoutput{\textbf{Output:~}}
  \algnewcommand\algorithmicparameters{\textbf{Parameters:~}}
  \algnewcommand\Input{\State\algorithmicinput}%
  \algnewcommand\Output{\State\algorithmicoutput}%
  \algnewcommand\Parameters{\State\algorithmicparameters}%
  \algnewcommand{\LineComment}[1]{\State \(\triangleright\) #1}

  \caption{BFS flood-fill algorithm for cluster classification.}\label{alg:floodfill}

  \begin{algorithmic}[1]
    \footnotesize

    \Input

    \Indent

    \State $\mathcal{C} = \left\{ \pnt{p}_1, \pnt{p}_2 \ldots \pnt{p}_N \right\}$ \Comment{a cluster of points}

    \State $\map$ \Comment{the current voxel grid map}

    \EndIndent

    \Output

    \Indent

    \State $\mathrm{floating} \in \left\{ \text{true}, \text{false} \right\}$ \Comment{whether the cluster is surrounded by air}

    \EndIndent

    \Parameters

    \Indent

    \State $d_{\text{search}} \in \mathbb{N}$ \Comment{the maximal search distance (in voxels)}

    \State $\mapval_{\text{unc}}, ~\mapval_{\text{tent}}$ \Comment{the \textit{uncertain} and \textit{tentative occupied} state thresholds}

    \EndIndent

    \For{each $ \pnt{p} \in \mathcal{C} $}

      \State $\vox_0 \coloneqq \vox(\pnt{p})$ \Comment{the starting voxel corresponding to $\pnt{p}$}

      \State $\mathcal{V}_{\text{explored}} \coloneqq \emptyset$ \Comment{a set of voxels explored so far}

      \State $\mathcal{V}_{\text{queue}} \coloneqq \left\{ \vox_0 \right\}$ \Comment{a FIFO queue of voxels to be explored}

      \While{$ \mathcal{V}_{\text{queue}} \neq \emptyset $}

        \State $\vox \coloneqq \text{pop}\left( \mathcal{V}_{\text{queue}} \right)$ \Comment{take the first element from the queue}

        \LineComment{if at least a \textit{tentative occupied} voxel is reached, terminate}
        \If { $ \mapval(\vox) \geq \mapval_{\text{tent}} $ } \label{alg:floodfill:if1start}

          \State \textbf{return} $\text{floating} \coloneqq \text{false}$

        \EndIf \label{alg:floodfill:if1end}
        
        \LineComment{if the voxel is \textit{uncertain}, check it}
        \If { $ \mapval(\vox) \in \interval[open right]{\mapval_{\text{unc}}}{\mapval_{\text{tent}}} $}

          \LineComment{if a path to a border of the search sphere is found, terminate}
          \If { $ \norm{ \pnt{c}_\vox - \pnt{c}_{\vox_0} } \geq d_{\text{search}} $ } \label{alg:floodfill:if2start}

            \State \textbf{return} $\text{floating} \coloneqq \text{false}$

          \EndIf \label{alg:floodfill:if2end}

          \LineComment{otherwise, expand the current voxel}
          \For{each $ \vox_{\text{neigh}} $ in a 6-neighborhood of $\vox$}

            \If { $\vox_{\text{neigh}} \notin \mathcal{V}_{\text{explored}} \land \vox_{\text{neigh}} \in \map$ }

              \State $ \text{push}\left( \vox_{\text{neigh}}, \mathcal{V}_{\text{queue}} \right) $ \Comment{add the element to the end}

            \EndIf

          \EndFor

        \EndIf \Comment{\textit{confident free} voxels are not expanded}

        \State $ \mathcal{V}_{\text{explored}} \coloneqq \mathcal{V}_{\text{explored}} \cup \vox $ \Comment{add $\vox$ to the explored set}

      \EndWhile

    \EndFor

    \State \textbf{return} $\text{floating} \coloneqq \text{true}$ \Comment{if all points passed, the cluster is floating}

  \end{algorithmic}

\end{algorithm}
}


Let $\mathcal{P}_{\text{bg}}$ be the set of all points from clusters classified as \textbf{background}, $\mathcal{P}_{\text{det}}$ be the set of points corresponding to \textbf{flying objects}, and $\mathcal{P}_{\text{unk}}$ be the set of \textbf{unknown} points where $\mathcal{P}_{\text{bg}} \cap \mathcal{P}_{\text{det}} = \mathcal{P}_{\text{bg}} \cap \mathcal{P}_{\text{unk}} = \mathcal{P}_{\text{det}} \cap \mathcal{P}_{\text{unk}} = \emptyset$.
Then, the output of the clustering and classification algorithm is
\begin{align}
  \mathcal{V}_{\text{bg}} &= \left\{ \vox(\pnt{p}) \mid \pnt{p} \in \mathcal{P}_{\text{bg}} \right\}, &&\mapw_{\text{bg}}(\vox) = \abs{ \mathcal{P}_{\text{bg}} \cap \vox },\\
  \mathcal{V}_{\text{det}} &= \left\{ \vox(\pnt{p}) \mid \pnt{p} \in \mathcal{P}_{\text{det}} \right\}, &&\mapw_{\text{det}}(\vox) = \infty,\\
  \mathcal{V}_{\text{unk}} &= \left\{ \vox(\pnt{p}) \mid \pnt{p} \in \mathcal{P}_{\text{unk}} \right\}, &&\mapw_{\text{unk}}(\vox) = \abs{ \mathcal{P}_{\text{unk}} \cap \vox }.
\end{align}
Note that $\mapw_{\text{det}}$ is infinity, so the corresponding voxels are simply set to $\mapfun_{\text{unk}}$.
Furthermore, for each cluster $\set{C}_D$ classified as a flying object, the position $\pnt{r}_D$ of the detection $D$ is calculated as
\begin{equation}
  \pnt{r}_D = \pnt{c}\left( \set{C}_D \right) \equiv \frac{1}{ \abs{\set{C}_D} } \sum_{\pnt{p} \in \set{C}_D} \pnt{p}. \label{eq:detpos}
\end{equation}


\begin{figure*}
  \centering
  \begin{subfigure}[t]{0.33\textwidth}
    \centering
    \includegraphics[width=\textwidth]{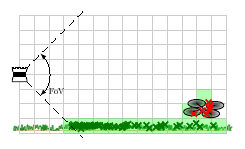}
    \caption{%
      Situation at time $t_0$.
    }
    \label{fig:sepbg_t1}
  \end{subfigure}%
  ~
  \begin{subfigure}[t]{0.33\textwidth}
    \centering
    \includegraphics[width=\textwidth]{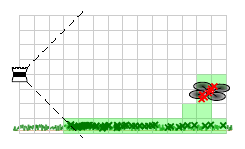}
    \caption{%
      Situation at time $t_1$.
    }
    \label{fig:sepbg_t2}
  \end{subfigure}%
  ~
  \begin{subfigure}[t]{0.33\textwidth}
    \centering
    \includegraphics[width=\textwidth]{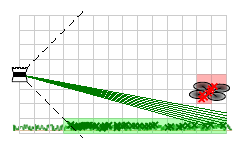}
    \caption{%
      Situation after raycasting (relevant rays are marked).
    }
    \label{fig:sepbg_t3}
  \end{subfigure}%
  \caption{%
    Illustration of a slow-moving object causing some areas of the map to be misclassified as occupied by traditional mapping algorithms.
    This problem can also manifest with the algorithm presented in this paper for a \ac{UAV} taking off.
    Points corresponding to the \ac{UAV} (marked with red crosses) are clustered with the background points (marked green) when it is landed (\ref{fig:sepbg_t1}), so the corresponding voxels are updated as occupied (green squares).
    When the \ac{UAV} takes off, points of the \ac{UAV} are still close to occupied voxels and the situation repeats (\ref{fig:sepbg_t2}).
    Raycasting the new points will correctly clear out the voxels previously occupied by the \ac{UAV}, but voxels currently containing points of the \ac{UAV} will still be misclassified (red squares in \ref{fig:sepbg_t3}).
    This problem is addressed by the algorithm described in section~\ref{sec:sepbg}.
  }
  \label{fig:sepbg}
\end{figure*}


\subsection{Raycasting}
\label{sec:raycasting}
A generalization of the raycasting algorithm presented in~\cite{amanatides1987fast} to three dimensions is used to find the set $\set{\vox}_{\text{int}}$ containing voxels intersected by the LiDAR's rays and the corresponding update weights $\mapw_{\text{int}}(\vox)$.
A set of line segments $\set{R}$ corresponding to the LiDAR's rays is obtained from the point cloud $\set{P}$.
If $\set{P}$ is an \textit{organized point cloud}, it also contains points corresponding to rays that did not hit an object within the sensor's range, which are typically set to some reserved value $\pnt{p}_{\text{empty}}$ to keep the organized structure.
Although these elements do not carry range information, they are important for updating empty voxels in $\map$.
However, due to the nature of most \acp{LiDAR}, many sensors do not distinguish between no hits and ray hits below the sensor's minimal range.
Assuming a sensor onboard a \ac{UAV} flying sufficiently far from obstacles, ray hits below the minimal range correspond to the body of the \ac{UAV}.
If the sensor is statically mounted, the order of these rays in an organized point cloud does not change.
We filter out such rays using a manually created mask $\set{I}_{\text{mask}}$ of the corresponding elements $\pnt{p}_i \in \set{P},~i \in \set{I}_{\text{mask}}$, which are ignored.

The remaining points in the point cloud $\set{P}$ are converted to line segments $\set{R}$.
For each element $\pnt{p}_i \in \set{P}, ~i\notin \set{I}_{\text{mask}}$, a line segment $r_i$ between the sensor's center $\pnt{c}_{\text{sens}}$ and an endpoint $\pnt{e}_i$ is added to $\set{R}$ with 
\begin{equation}
  \pnt{e}_i = \begin{cases}
    \pnt{c}_{\text{sens}} + d_{\text{max}} \vec{d}_i,  & \text{if } \pnt{p}_i = \pnt{p}_{\text{empty}}, \\
    \pnt{c}_{\text{sens}} + \min\left( d_{\text{max}},\, \norm{\pnt{p}_i - \pnt{c}_{\text{sens}}} \right) \vec{d}_i,  & \text{else,}
  \end{cases}
\end{equation}
where $\vec{d}_i$ is a direction vector of the sensor's $i$-th ray and $d_{\text{max}}$ is a parameter of the algorithm.

Each line segment $r \in \set{R}$ is processed by the raycasting algorithm, which provides a length $l_{\text{int}}(\vox, r)$ of the section of $r$ contained within every intersected voxel $\vox$.
These are accumulated per voxel to obtain the update weights $\mapw_{\text{int}}(\vox)$.
The maximal amount of information a ray can provide about a voxel's occupancy is achieved when their intersection is the longest, which corresponds to the case when the ray goes through the voxel's diagonal.
Based on this intuition, we normalize the accumulated lengths by the length of the voxel diagonal $\sqrt{3}s_v$ to obtain the equivalent number of such most informative intersections.

In order not to delay the output of the detections, the raycasting runs in parallel to the clustering \& classification algorithm described in the previous section.
However, this may cause problems when one of the algorithms processes data at a faster rate, resulting in an unbalanced updating of the map.
To compensate for this, the update weighting coefficient of the slower algorithm is multiplied by the corresponding number of updates of the faster algorithm.
Typically, the clustering \& classification is the fastest, so we formally define the compensation here.
The output of the raycasting algorithm is then
\begin{align}
  \mathcal{V}_{\text{int}} &= \left\{ \vox \in \map~\mid~\exists\,r \in \set{R},~l_{\text{int}}(\vox, r) \neq 0 \right\},\\
  \mapw_{\text{int}}(\vox) &= N_{\text{cls}} w_{\text{int}} \frac{\sqrt{3}}{3 s_v}\sum_{r\in\set{R}} l_{\text{int}}(\vox, r),
\end{align}
where $N_{\text{cls}}$ is the number of the clustering and classification updates from the start of raycasting, $w_{\text{int}}$ is a weighting factor used to tune the aggressiveness of the raycasting, and $s_v$ is size (edge length) of the voxels.



\subsection{Separate background voxel removal}
\label{sec:sepbg}

This algorithm runs in parallel to the Raycasting and Clustering \& classification modules (see Fig.~\ref{fig:schematic_overview}) and serves to reset the \textit{tentative occupied} voxel clusters in $\set{M}$ that are separated from the \textit{confident occupied} clusters.
The main purpose of this algorithm is to improve the detection of a target that has taken off from the ground.
In a traditional occupancy mapping method, a slow-moving object leaves a trail of occupied voxels and the points corresponding to the target are also classified as occupied (see Fig.~\ref{fig:graph_update_rule}).
Because the mapping method presented in this paper only updates voxels as occupied if they contain points close to other occupied voxels (as explained in the previous sections), this problem is largely mitigated.
However, this method can fail if an object is closer to the background than $d_{\text{close}}$ and then leaves, such as during a \ac{UAV} take-off.
This module addresses this scenario, as illustrated in Fig.~\ref{fig:sepbg}.

Firstly, voxels from $\set{M}$ with $\mapval(\vox) \geq \mapval_{\text{tent}}$ are separated into Euclidean clusters $\set{\vox}_{\text{cl},i}$ using the same method as in section~\ref{sec:clustering} with a minimal inter-cluster distance $d_{\text{sep}}$.
For each voxel cluster $\set{\vox}_{\text{cl},i}$, the number of voxels that are \textit{confident occupied} is counted:
\begin{equation}
  N_{\text{conf},i} = \abs{ \left\{ \vox \in \set{\vox}_{\text{cl},i} \mid \mapval(\vox) \geq \mapval_{\text{conf}} \right\} }.
\end{equation}
The output of this algorithm is then
\begin{align}
  \mathcal{V}_{\text{sep}} &= \bigcup_{ i \in \set{I}_{\text{sep}} }\set{V}_{\text{cl},i}, ~ \set{I}_{\text{sep}} = \left\{ i \mid N_{\text{conf},i} < N_{\text{conf,min}} \right\} \\
  \mapw_{\text{int}}(\vox) &= N_{\text{cls}},
\end{align}
where $N_{\text{conf,min}}$ is a parameter of the algorithm, and $N_{\text{cls}}$ is the number of parallel updates by the clustering \& classification algorithm, as discussed in the previous section.


\subsection{A priori map initialization}
In a general unknown environment, all voxels in $\set{M}$ are initialized to $\mapfun_{\text{unk}}$, corresponding to the \textit{unknown} occupancy state.
However, an a priori knowledge of the environment can be utilized to initialize the map simply by setting the pre-mapped voxels to the respective values.
If changing the states of the pre-mapped voxels is undesired, the values of voxels known to be occupied a priori can be set to $\infty$ and values of a priori free voxels to $-\infty$.
This ensures that the state of these voxels remains unchanged by the algorithms described thus far and that the voxels are always classified as \textit{confident occupied} and \textit{confident free}, respectively.



\subsection{LiDAR-based multi-target tracking}

The proposed multi-target tracking algorithm is shown in Alg.~\ref{alg:mtt}.
The inputs of the tracker are point clouds $\set{P}$, detections $\set{D}$, and occupied voxels $\set{O}$ in the map $\set{M}$, which are all provided by the modules described above.
The algorithm also relies on time-stamps $t\tstep{k}$ of the input data to accurately predict states of the tracks (see the \ac{KF} model description in sec.~\ref{sec:kf}).
The tracker keeps a \ac{FIFO} buffer $\set{B}$ of the latest $N_{\text{buf}}$ point clouds ordered by their time-stamps and a set of active tracks $\set{T}$.
The algorithm is updated whenever a new point cloud is received using the $\mathrm{newPointCloud}()$ routine, and when a new set of detections is received using $\mathrm{newDetections}()$.
Both these routines rely on the function $\mathrm{updateTrack}()$ for state prediction, track-to-measurement association, and state correction using the associated measurement and a \ac{KF} model.

\subsubsection{Track update}
\label{sec:kf}


{
\begin{algorithm}
  \algdef{SE}[SUBALG]{Indent}{EndIndent}{}{\algorithmicend\ }%
  \algtext*{Indent}
  \algtext*{EndIndent}

  \algnewcommand\AND{\textbf{and}~}
  \algnewcommand\Not{\textbf{not}~}
  \algnewcommand\Or{\textbf{or}~}
  \algnewcommand\algorithmicinput{\textbf{Input:~}}
  \algnewcommand\algorithmicoutput{\textbf{Output:~}}
  \algnewcommand\algorithmicparameters{\textbf{Parameters:~}}
  \algnewcommand\Input{\State\algorithmicinput}%
  \algnewcommand\Output{\State\algorithmicoutput}%
  \algnewcommand\Parameters{\State\algorithmicparameters}%
  \algnewcommand\Routine{\vspace{0.5em}\State\textbf{Routine~}}%
  \algnewcommand\InnerState{\State\textbf{Persistent state:~}}%
  \algnewcommand{\LineComment}[1]{\State \(\triangleright\) #1}

  \caption{Point cloud multi-target tracking algorithm}\label{alg:mtt}

  \begin{algorithmic}[1]
    \footnotesize

    \Input

    \Indent

    \State $\set{P}\tstep{k} = \left\{ \pnt{p}_i \right\}$ \Comment{a scan from the \ac{LiDAR} at time-step $k$}

    \State $\set{D}\tstep{k} = \left\{ D_i \right\}$ \Comment{a set of detections at time-step $k$}

    \State $\mathcal{O} = \left\{ \pnt{c}\left(\vox\right) ~\mid~ \vox \in \set{M},~ \mapval(\vox) \geq \mapval_{\text{tent}} \right\}$ \Comment{occupied voxels' centers}

    \State $t\tstep{k} \in \mathbb{R}$ \Comment{time of time-step $k$}

    \EndIndent

    \InnerState

    \Indent

    \State $\set{T} = \left\{ T_i \right\}$ \Comment{set of active tracks where $T \equiv \left\{ \hat{\pnt{x}}_{T}, \mat{P}_{T} \right\}$}

    \State $\set{B} = \left\{ \set{P}\tstep{i} \right\}$ \Comment{buffer of the last $N_{\text{buf}}$ point clouds}

    \EndIndent

    \Routine $\mathrm{newPointCloud}\left( \set{P}\tstep{k},~\set{O} \right)$:

    \Indent

      \State $\mathrm{push}\left( \set{P}\tstep{k},~\set{B} \right)$ \Comment{add $\set{P}\tstep{k}$ to the front of the buffer} \label{alg:mtt:bufstart}

      \If { $ \abs{ \set{B} } > N_{\text{buf}} $ }

        \State $\mathrm{pop}\left( \set{B} \right)$ \Comment{if the buffer is full, remove the last element}

      \EndIf \label{alg:mtt:bufend}

      \For{each $ T \in \set{T} $} \label{alg:mtt:track1start}

        \State $T \coloneqq \mathrm{updateTrack}\left( T,~\set{P}\tstep{k},~\set{O},~t\tstep{k}-t\tstep{k-1} \right)$

        \If{$ r_{\text{unc}}\left( \mat{P}_{T} \right) > r_{\text{max}}$}

          \State $\set{T} \coloneqq \set{T} \setminus T$ \Comment{remove tracks that are too uncertain}

        \EndIf

      \EndFor \label{alg:mtt:track1end}

      \rone{\For{each $ \left( T_i, T_j \right) \in \set{T} \times \set{T} $} \label{alg:mtt:remtracksstart} \Comment{iterate over all track combinations}

        \If{$ i \neq j $ \textbf{and} $ \norm{ \hat{\pnt{r}}_{T_i} - \hat{\pnt{r}}_{T_j} } < r_{\text{sel}}\left( \mat{P}_{T_i} \right) +  r_{\text{sel}}\left( \mat{P}_{T_j} \right)$}

          \If{$ r_{\text{unc}}\left( \mat{P}_{T_i} \right) > r_{\text{unc}}\left( \mat{P}_{T_j} \right)$}

            \State $\set{T} \coloneqq \set{T} \setminus T_i$ \Comment{remove the more uncertain track}

          \Else

            \State $\set{T} \coloneqq \set{T} \setminus T_j$ \Comment{remove the more uncertain track}

          \EndIf

        \EndIf

      \EndFor \label{alg:mtt:remtracksend}}

    \EndIndent

    \Routine $\mathrm{newDetections}\left( \set{D}\tstep{k},~\set{O} \right)$:

    \Indent

      \For{each $ D \in \set{D}\tstep{k} $}

        \State $T^* \coloneqq \mathrm{initializeTrack}\left( D \right)$ \label{alg:mtt:trinit} \Comment{described in eqs.~\eqref{eq:track_init1}-\eqref{eq:track_init2}}

        \For{each $ \set{P}\tstep{i} \in \set{B},~t\tstep{i} > t\tstep{k} $} \label{alg:mtt:detupstart} \Comment{propagate $T^*$ through $\set{B}$}

          \State $T^* \coloneqq \mathrm{updateTrack}\left( T^*,~\set{P}\tstep{i},~\set{O},~t\tstep{i}-t\tstep{i-1} \right)$

        \EndFor \label{alg:mtt:detupend}

        \If{$ r_{\text{unc}}\left( \mat{P}_{T^*} \right) \leq r_{\text{max}}$} \label{alg:mtt:detaddstart} \Comment{ignore $T^*$ if it is too uncertain}


          \If{$ \exists T \in \set{T}, ~ \norm{ \hat{\pnt{r}}_{T} - \hat{\pnt{r}}_{T^*} } \leq r_{\text{unc}}\left( \mat{P}_{T} \right) + r_{\text{unc}}\left( \mat{P}_{T^*} \right) $}

            \State $n_{\text{det}}(T) \coloneqq n_{\text{det}}(T) + 1$ \Comment{update the similar track}

          \Else

            \State $\set{T} \coloneqq \set{T} \cup T^*$ \Comment{or add $T^*$ as a new track to $\set{T}$}

          \EndIf

        \EndIf \label{alg:mtt:detaddend}

      \EndFor

    \EndIndent

    \Routine $\mathrm{updateTrack}\left( T,~\set{P},~\set{O},~{}_\Delta t \right)$: \label{alg:mtt:updateTrackstart}

    \Indent

    \State $T \coloneqq \mathrm{predictKF}\left( T,~{}_\Delta t \right)$ \label{alg:mtt:predict}

    \LineComment{select points within $r_{\text{sel}}$ of the position estimate $\hat{\pnt{r}}_T$}

    \State $\set{P}' \coloneqq \left\{ \pnt{p} \in \set{P} ~\mid~ \norm{ \pnt{p} - \hat{\pnt{r}}_{T} } \leq r_{\text{sel}}\left( \mat{P}_T  \right) \right\}$\label{alg:mtt:measstart}

    \State $\set{K} \coloneqq \mathrm{extractClusters}\left( \set{P}' \right)$

    \LineComment{filter out clusters too close to occupied voxels $\set{O}$}

    \State $\set{K}' \coloneqq \left\{ \set{C} \in \set{K}~\mid~ \min_{v \in \set{O}}\left( \norm{ \pnt{c}\left(\set{C}\right) - \pnt{c}\left(\vox\right) } \right) > d_{\text{min}} \right\}$ \label{alg:mtt:bgrem}

    \If{ $\set{K}' \neq \emptyset$ }

      \LineComment{use the centroid of the closest cluster for KF correction}

      \State $\set{C}^* := \argmin_{\set{C} \in \set{K}'} \norm{ \pnt{c}\left( \set{C} \right) - \hat{\pnt{r}}_{T} }$

      \State $\pnt{z} := \pnt{c}\left({\set{C}^*}\right) $\label{alg:mtt:measend}

      \State $T \coloneqq \mathrm{correctKF}\left( T,~\pnt{z} \right)$ \label{alg:mtt:correct}

    \EndIf

    \State \textbf{return} $T$ \label{alg:mtt:updateTrackend}

    \EndIndent

  \end{algorithmic}

\end{algorithm}
}


The $\mathrm{updateTrack}()$ subroutine (lines~\ref{alg:mtt:updateTrackstart}-\ref{alg:mtt:updateTrackend} of Alg.~\ref{alg:mtt}) updates a track $T$, which consists of a state estimate $\hat{\pnt{x}}_T$, its corresponding covariance matrix $\mat{P}_T$, and the number of associated detections $n_{\text{det}}(T)$.
The state estimate
\begin{equation}
  \hat{\pnt{x}}_T = \bemat{
    \hat{\pnt{r}}_T\tran &\hat{\vec{v}}_T\tran &\hat{\vec{a}}_T\tran
  }\tran
\end{equation}
approximates the tracked object's state vector
\begin{equation}
  \pnt{x} = \bemat{
    \pnt{r}\tran &\vec{v}\tran &\vec{a}\tran
  }\tran,
\end{equation}
where $\pnt{r} \in \mathbb{R}^3$ is the position of the object, $\vec{v} \in \mathbb{R}^3$ is its velocity, and $\vec{a} \in \mathbb{R}^3$ is the acceleration.
The covariance matrix $\mat{P}_T$ represents the uncertainty of the estimate as a Gaussian distribution.
$n_{\text{det}}(T)$ can be used to reject sporadic false positives or to select the primary track in some applications.

Firstly, the estimate of the state and its uncertainty are propagated forward by ${}_\Delta t\tstep{k} = t\tstep{k} - t\tstep{k-1}$ to the time $t\tstep{k}$ of acquisition of the input point cloud $\set{P}$, where $t\tstep{k-1}$ is the time of the last update of $T$.
Then, points $\set{P}'$ within the search radius $r_{\text{sel}}$ of the track are selected (line~\ref{alg:mtt:measstart}).
The radius $r_{\text{sel}}$ is scaled with the track's covariance $\mat{P}_T$ as
\begin{align}
  r_{\text{sel}}\left( \mat{P}_T \right) &= \max\left( r_{\text{min}},~r_{\text{unc}}\left( \mat{P}_T \right) \right), \\
  r_{\text{unc}}\left( \mat{P}_T \right) &= c_r \sqrt[3]{\det\left( \mat{H} \mat{P}_T \mat{H}\tran \right) } , \label{eq:r_unc}
\end{align}
where $r_{\text{min}}$ is a minimal search radius parameter, $c_r$ specifies a confidence interval for a Gaussian distribution, $\mat{H}$ is a position measurement matrix defined in \autoref{eq:measmat},
and $r_{\text{unc}}\left( \mat{P}_T \right)$ can be interpreted as a radius of a sphere with a volume equivalent to the $c_r$-confidence ellipsoid of the predicted position $\hat{\pnt{r}}_T$.
This method is used as a compromise between accurate representation of the uncertainty and an efficient implementation of the radius search using a KD-tree.
The selected points $\set{P}'$ are separated into clusters $\set{K}$ using the same method as in sec.~\ref{sec:clustering}.
These clusters are filtered into $\set{K}'$ based on their distance to the nearest occupied voxel from $\set{O}$ (line~\ref{alg:mtt:bgrem}) in order to remove clusters corresponding to background objects.
Finally, if the filtered set $\set{K}'$ is not empty, the cluster $\set{C}$ with the centroid $\pnt{c}\left( \set{C} \right)$ closest to the track's predicted position $\hat{\pnt{r}}_T$ is selected.
The centroid is then used as a \ac{KF} measurement $\pnt{z}$ to correct the track's state estimate and covariance.

To update a track $T$ using a measurement $\pnt{z}$ and to predict the tracked object's future trajectory, a linear \ac{KF} is used.
Motion of each object is modeled as a point mass with second order dynamics. Perturbances of the system are modeled as Gaussian noise.
We use a standard discrete state-space mathematical representation of the model with the state vector $\pnt{x}$ defined above.
The state-transition matrix is
\begin{equation}
  \mat{A}\tstep{k} = \bemat{
    \mat{I} & {}_\Delta t\tstep{k} \mat{I} & \frac{1}{2}{}_\Delta t\tstep{k}^2\mat{I} \\
    \mat{0} & \mat{I} & {}_\Delta t\tstep{k} \mat{I} \\
    \mat{0} & \mat{0} & \mat{I} \\
  },
\end{equation}
where $\mat{I} \in \mathbb{R}^{3\times 3}$ is an identity matrix, $\mat{0} \in \mathbb{R}^{3\times 3}$ is a zero matrix, and ${}_\Delta t\tstep{k}$ is the duration since the previous time step $k-1$.
The state-space model of the track's motion is then
\begin{align}
  \pnt{x}\tstep{k+1} = \mat{A}\tstep{k}\pnt{x}\tstep{k} + \pnt{\xi}\tstep{k}, && \pnt{\xi}\tstep{k} \sim \mathcal{N}\left(\pnt{0}, \mat{\Xi}\tstep{k}\right), \label{eq:ss}
\end{align}
where $\pnt{\xi}$ is Gaussian noise with zero mean and a covariance matrix $\mat{\Xi}$.
Similarly, the measurement model is
\begin{align}
  \pnt{z}\tstep{k} = \mat{H}\pnt{x}\tstep{k} + \pnt{\zeta}\tstep{k}, && \pnt{\zeta}\tstep{k} \sim \mathcal{N}\left(\pnt{0}, \mat{Z}\tstep{k}\right). \label{eq:meas}
\end{align}
It is assumed that the cluster centroid $\pnt{c}\left(\set{C}^*\right)$ used as the measurement $\pnt{z}$ corresponds to the position of the tracked object.
The measurement matrix $\mat{H}$ is therefore defined as
\begin{equation}
  \mat{H} = \bemat{
    \mat{I} & \mat{0} & \mat{0}
  }. \label{eq:measmat}
\end{equation}
The process noise covariance $\mat{\Xi}\tstep{k}$ was empirically identified using parameters $\Xi_{\pnt{r}}$, $\Xi_{\pnt{v}}$, and $\Xi_{\pnt{a}}$ as
\begin{equation}
  \mat{\Xi}\tstep{k} = \mat{\Xi} = \bemat{
    \Xi_{\pnt{r}}^2 \mat{I} & \mat{0} & \mat{0} \\
    \mat{0} & \Xi_{\pnt{v}}^2 \mat{I} & \mat{0} \\
    \mat{0} & \mat{0} & \Xi_{\pnt{a}}^2 \mat{I}
  }
\end{equation}
The measurement noise covariance $\mat{Z}\tstep{k}$ may be selected similarly using a parameter $Z$ as $\mat{Z}\tstep{k} = \mat{Z} = Z^2\mat{I}$ or based on known uncertainties of the sensor's pose as discussed in sec.~\ref{sec:meas_cov}.

Equations~(\ref{eq:ss}) and~(\ref{eq:meas}) are used by the \ac{KF} to implement the prediction and correction steps which calculate the state estimate $\hat{\pnt{x}}_T$ and its covariance matrix $\mat{P}_T$ based on the time step ${}_\Delta t$ and measurement $\pnt{z}$ (lines~\ref{alg:mtt:predict} and~\ref{alg:mtt:correct}).

\subsubsection{Point cloud update}
The $\mathrm{newPointCloud}()$ routine processes incoming \ac{LiDAR} scans $\set{P}$.
These are used to update the buffer $\set{B}$ (lines~\ref{alg:mtt:bufstart}-\ref{alg:mtt:bufend}) and the set of active tracks $\set{T}$ (lines~\ref{alg:mtt:track1start}-\ref{alg:mtt:track1end}).
We assume that some tracked objects can disappear (e.g. landed or crashed \acp{UAV}) and that the detector can produce sporadic false positives.
Such objects will no longer be represented in the point clouds with non-background points and the corresponding tracks will therefore not be corrected in the $\mathrm{updateTrack}()$ routine.
The uncertainty of such tracks will grow without bounds since the \ac{KF} prediction step is still applied.
To filter these out, the tracks whose position uncertainty radius $r_{\text{unc}}$ grows beyond a threshold parameter $r_{\text{max}}$ are removed from $\set{T}$.
\rone{Furthermore, track duplicates that can emerge e.g. due to false positives are removed (lines~\ref{alg:mtt:remtracksstart}-\ref{alg:mtt:remtracksend}).
The duplicates are detected by comparing the mutual distance of the two tracks to a dynamic threshold determined by the corresponding search radii.}

\subsubsection{Detections update}
Each detection $D$ from a newly received set of detections $\set{D}$ initializes a new tentative track $T^*$ (line \ref{alg:mtt:trinit}) with a state estimate and covariance matrix
\begin{align}
  \hat{\pnt{x}} &\coloneqq \bemat{
    \pnt{r}_{D}\tran & \pnt{0} & \pnt{0}
  }\tran, \label{eq:track_init1} \\
  \mat{P} &\coloneqq \bemat{
    P_{\text{0},\pnt{r}}^2 \mat{I} & \mat{0} & \mat{0} \\
    \mat{0} & P_{\text{0},\pnt{v}}^2 \mat{I} & \mat{0} \\
    \mat{0} & \mat{0} & P_{\text{0},\pnt{a}}^2 \mat{I}
  },\label{eq:track_init2}
\end{align}
where $\pnt{r}_D$ is the position of the detection (obtained in eq.~\ref{eq:detpos}) and $P_{\text{0},\pnt{r}}$, $P_{\text{0},\pnt{v}}$, $P_{\text{0},\pnt{a}}$ are empirically determined parameters.
Note that similarly as for eq.~\eqref{eq:meas}, the empirical initialization of the submatrix of $\mat{P}$ corresponding to position can be replaced by a more accurate solution in case the pose uncertainty of the sensor is known, as discussed in sec.~\ref{sec:meas_cov}.

This new track $T^*$ is sequentially updated using point clouds from $\set{B}$ that are newer than the detection time $t\tstep{k}$ (lines \ref{alg:mtt:detupstart}-\ref{alg:mtt:detupend}).
If the uncertainty radius $r_{\text{unc}}\left( \mat{P}_{T^*} \right)$ (defined in eq.~\eqref{eq:r_unc}) of the updated track $T^*$ is higher than a threshold $r_{\text{max}}$, it is considered either lost or a false positive and is discarded.
Otherwise, if a similar track $T \in \set{T}$ already exists, its number of associated detections $n_{\text{det}}(T)$ is incremented.
If $T^*$ is not discarded and no similar track exists, it is added to the set of active tracks $\set{T}$ (lines \ref{alg:mtt:detaddstart}-\ref{alg:mtt:detaddend}).



\section{Theoretical Analysis}
\label{sec:theoretical_analysis}

In this section, we present an analysis of probabilistic properties of measurements by a \ac{LiDAR} or a similar sensor with uncertainty in the pose of the sensor and the measured range.
The analysis is used to estimate the expected Euclidean error of a measured point and to evaluate the probability of updating the correct voxel in voxel-based mapping.
The results have practical applications, such as evaluation of the limitations of occupancy mapping accuracy with respect to known uncertainties, more accurate fusion of the measurements using methods that can take into account covariance of the measurements (e.g. the \ac{KF}), or predicting the position error of a detected target.
These can be used within the system presented in this paper, as well as in related problems.


\subsection{LiDAR measurement uncertainty}
\label{sec:meas_cov}

Let us consider a single point $\pnt{p}_m$ measured by a ray of a \ac{LiDAR} sensor with a (unit) direction vector $\vec{d}$ and a range $l_m$.
The pose of the sensor in a static world frame $\mathcal{W}$ is measured as a translation vector $\pnt{t}_m$ and rotation matrix $\mat{R}_m$.
The measured point $\pnt{p}_m$ can be expressed as a function of the measured range, translation, and rotation as
\begin{align}
  \pnt{p}_m &= l_m \mat{R}_{m} \vec{d} + \pnt{t}_{m}. \label{eq:meas_pt}
\end{align}
If the ground truth of the sensor's pose and of the range measurement were known with absolute accuracy, the corresponding ground-truth point $\pnt{p}_{gt}$ could be obtained as
\begin{align}
  \pnt{p}_{gt} &= l_{gt}\mat{R}_{gt}\vec{d} + \pnt{t}_{gt}, \label{eq:gt_pt}
\end{align}
where $l_{gt}$, $\pnt{t}_{gt}$ and $\mat{R}_{gt}$ are the ground-truth (noiseless) range and pose.

We model the relation between the measured values $l_m$, $\pnt{t}_m$, $\mat{R}_m$ and the corresponding ground truth as
\begin{align}
  l_{gt} &= l_{m} + l_n, \\
  \pnt{t}_{gt} &= \pnt{t}_{m} + \pnt{t}_n, \label{eq:pose_meas_t} \\
  \mat{R}_{gt} &= \mat{R}_{m} \mat{R}_{n}
  = \mat{R}_{m} \mat{R}_{z}\left( \gamma_n \right) \mat{R}_{y}\left( \beta_n \right) \mat{R}_{x}\left( \alpha_n \right), \label{eq:pose_meas_l}
\end{align}
where $l_n$, $\pnt{t}_n$ and $\alpha_n$, $\beta_n$, $\gamma_n$ represent the unknown measurement noise.
Let us define a noise vector $\pnt{w}$ that is assumed to be drawn from a multivariate Gaussian distribution as
\begin{align}
  \pnt{w} &= \bemat{ l_n & \pnt{t}_n\tran & \pnt{a}_n\tran }\tran, \\
  \pnt{w} &\sim \mathcal{N}\left( \pnt{0}, \mat{\Sigma}_{\pnt{w}} \right), \label{eq:noise_vec}
\end{align}
where $\pnt{a}_n = \bemat{ \alpha_n, \beta_n, \gamma_n }\tran$ and $\mat{\Sigma}_{\pnt{w}}$ is a known covariance matrix of the measurement noise.

In practice, the true position of $\pnt{p}_{gt}$ is unknown, but its probability distribution can be estimated given the measured point $\pnt{p}_{m}$ and the measurement uncertainty $\mat{\Sigma}_{\pnt{w}}$.
Although it does not have a practical analytical solution, the probability distribution can be approximated from eqs.~\eqref{eq:gt_pt}-\eqref{eq:pose_meas_l} using linearization as a normal distribution with a mean $\pnt{\mu}$ and a covariance matrix $\mat{\Sigma}$.
The transformation of the known covariance $\mat{\Sigma}_{\pnt{w}}$ of the random variable $\pnt{w}$ to the covariance $\mat{\Sigma}$ is then derived as
\begin{align}
  \mat{\Sigma} = \mat{J} \mat{\Sigma}_{\pnt{w}} \mat{J}\tran, &&\mat{J} = \at{\frac{ \partial \pnt{p}_{gt}}{ \partial\pnt{w} }}{\pnt{w} = \pnt{0}}. \label{eq:pt_cov}
\end{align}
Using a substitution
{
\footnotesize
\begin{align}
  \dot{\mat{R}}_{\alpha_n} = \at{\frac{\partial \mat{R}_x }{\partial \alpha_n}}{\alpha_n = 0}, &&
  \dot{\mat{R}}_{\beta_n} = \at{\frac{\partial \mat{R}_y }{\partial \beta_n}}{\beta_n = 0}, &&
  \dot{\mat{R}}_{\gamma_n} = \at{\frac{\partial \mat{R}_z }{\partial \gamma_n}}{\gamma_n = 0},
\end{align}
}%
the Jacobian $\mat{J}$ evaluates to
{
\begin{align}
  &\mat{J} = \label{eq:jacobian}\\
  &\bemat{
    \mat{R}_{m}\vec{d}
  & \mat{I}
  & l_{m} \mat{R}_{m} \dot{\mat{R}}_{\alpha_n} \vec{d}
  & l_{m} \mat{R}_{m} \dot{\mat{R}}_{\beta_n} \vec{d}
  & l_{m} \mat{R}_{m} \dot{\mat{R}}_{\gamma_n} \vec{d}
  }. \nonumber
\end{align}
}%
The mean $\pnt{\mu}$ of the linearized distribution is the expected value of $\pnt{p}_{gt}$, which is obtained from eq.~\eqref{eq:gt_pt} as
\begin{equation}
  \pnt{\mu} = \expected{ \pnt{p}_{gt} } = \at{\pnt{p}_{gt}}{\pnt{w} = \pnt{0}} = \pnt{p}_m. \\
\end{equation}
The approximated probability density function of $\pnt{p}_{gt}$ is then
\begin{equation}
  f_{\pnt{p}_{gt}}\left( \pnt{p} \right) \approx f_{\mathcal{N}} \left( \pnt{p}, \pnt{\mu}, \mat{\Sigma} \right) = f_{\mathcal{N}} \left( \pnt{p}, \pnt{p}_m, \mat{J} \mat{\Sigma}_{\pnt{w}} \mat{J}\tran \right), \label{eq:approx_dist}
\end{equation}
which is a multivariate normal probability density function with mean $\pnt{\mu} = \pnt{p}_m$ and covariance $\mat{\Sigma} = \mat{J} \mat{\Sigma}_{\pnt{w}} \mat{J}\tran$.

This linearization provides a powerful tool for approximating the probability distribution of objects detected by a sensor with uncertain pose and distance estimate, which is a common problem in robotics that is, to our best knowledge, not sufficiently addressed in the literature.
Using this approximation, other properties of the problem can be quantified, as will be shown in the following sections.

The sensor model described by eqs.~\eqref{eq:meas_pt}-\eqref{eq:noise_vec} was simulated to obtain a \ac{MC} approximation of $f_{\pnt{p}_{gt}}\left( \pnt{p} \right)$ which was compared to the approximated distribution defined in eqs.~\eqref{eq:pt_cov}-\eqref{eq:approx_dist} for different parameters of the noise vector $\pnt{w}$.
To compare the probability distributions, an absolute probability difference $D_r$ over region $s$ was used, defined as
\begin{align}
  &D_r\left( s, \pnt{p}_{m}, \mat{\Sigma}_{\pnt{w}} \right) =\\
  &\hspace{4em} \abs{ p_{\text{mc}}\left( \pnt{p}_{gt} \in s \mid \pnt{p}_{m}, \mat{\Sigma}_{\pnt{w}} \right) - p_{\text{g}}\left( \pnt{p}_{gt} \in s \mid \pnt{p}_{m}, \mat{\Sigma}_{\pnt{w}} \right) } , \nonumber
\end{align}
where $p_{\text{mc}}\left( \pnt{p}_{gt} \in s \mid \pnt{p}_{m}, \mat{\Sigma}_{\pnt{w}} \right)$ is the empirical \ac{MC} probability that $\pnt{p}_{gt}$ lies within the region $s$ given a measured point $\pnt{p}_m$ and noise covariance $\mat{\Sigma}_{\pnt{w}}$, and $p_{\text{g}}\left( \pnt{p}_{gt} \in s \mid \pnt{p}_{m}, \mat{\Sigma}_{\pnt{w}} \right)$ is the corresponding approximated Gaussian probability.
The distributions were evaluated for all combinations of parameters $\sigma_{\si{\metre}}$, $\sigma_{\si{\radian}}$, and $d$ defined as
\begin{align}
  \mat{\Sigma}_{\pnt{w}} &= \diag{\left( \sigma_{\si{\metre}}, \sigma_{\si{\metre}}, \sigma_{\si{\metre}}, \sigma_{\si{\metre}}, \sigma_{\si{\radian}}, \sigma_{\si{\radian}}, \sigma_{\si{\radian}} \right)}^2, \\
  \sigma_{\si{\metre}} &\in \left\{ \num{1e-2}, \num{1e-1}, \num{1e-0} \right\},
  \hspace{0.5em} \sigma_{\si{\radian}} \in \left\{ \num{5e-3}, \num{5e-2}, \num{5e-1} \right\}, \\
  \vec{d} &= \bemat{1 & 0 & 0}\tran,
  \hspace{0.5em} l_{gt} = d, 
  \hspace{0.5em} \pnt{t}_{gt} = \pnt{0},
  \hspace{0.5em} \mat{R}_{gt} = \mat{R}_{\text{rand}}, \\
  d &\in \left\{ \SI{0}{\metre}, \SI{1}{\metre}, \dots, \SI{80}{\metre} \right\},
  \hspace{0.5em}\mat{R}_{\text{rand}} \sim \mathcal{U}\left(\text{SO}(3)\right),
\end{align}
where the rotation matrix $\mat{R}_{\text{rand}}$ was drawn from a uniform probability distribution over $\text{SO}(3)$ to marginalize out its influence on the distribution.
The ranges of the parameters were chosen based on our experiences with practical deployment of \acp{UAV} to represent different environmental conditions and self-localization methods of the \ac{UAV}.

\begin{table}
\footnotesize
\begin{tabularx}{\linewidth}{X X X | X X X}
\toprule
  $\sigma_{\si{\radian}}$ & $\sigma_{\si{\metre}}$ & $d$ & $\max\left( D_r \right)$ & $\bar{ D }_p$ & $\sigma_{ D_r }$ \\
\midrule
  $\SI{0.005}{\radian}$ & $\SI{0.01}{\metre}$ & $\SI{24}{\metre}$ & $\num{7.47e-05}$ & $\num{8.70e-07}$ & $\num{3.03e-06}$ \\
  $\SI{0.005}{\radian}$ & $\SI{0.01}{\metre}$ & $\SI{48}{\metre}$ & $\num{8.79e-05}$ & $\num{8.73e-07}$ & $\num{3.05e-06}$ \\
  $\SI{0.005}{\radian}$ & $\SI{0.01}{\metre}$ & $\SI{72}{\metre}$ & $\num{8.55e-05}$ & $\num{8.67e-07}$ & $\num{3.04e-06}$ \\
  $\SI{0.005}{\radian}$ & $\SI{0.1}{\metre}$  & $\SI{24}{\metre}$ & $\num{8.00e-05}$ & $\num{8.85e-07}$ & $\num{3.05e-06}$ \\
  $\SI{0.005}{\radian}$ & $\SI{0.1}{\metre}$  & $\SI{48}{\metre}$ & $\num{7.89e-05}$ & $\num{9.08e-07}$ & $\num{3.05e-06}$ \\
  $\SI{0.005}{\radian}$ & $\SI{0.1}{\metre}$  & $\SI{72}{\metre}$ & $\num{9.38e-05}$ & $\num{9.14e-07}$ & $\num{3.02e-06}$ \\
  $\SI{0.005}{\radian}$ & $\SI{1}{\metre}$    & $\SI{24}{\metre}$ & $\num{1.08e-04}$ & $\num{9.10e-07}$ & $\num{3.03e-06}$ \\
  $\SI{0.005}{\radian}$ & $\SI{1}{\metre}$    & $\SI{48}{\metre}$ & $\num{8.03e-05}$ & $\num{9.11e-07}$ & $\num{3.04e-06}$ \\
  $\SI{0.005}{\radian}$ & $\SI{1}{\metre}$    & $\SI{72}{\metre}$ & $\num{8.41e-05}$ & $\num{9.07e-07}$ & $\num{3.02e-06}$ \\
  $\SI{0.05}{\radian}$  & $\SI{0.01}{\metre}$ & $\SI{24}{\metre}$ & $\num{2.47e-04}$ & $\num{1.45e-06}$ & $\num{9.28e-06}$ \\
  $\SI{0.05}{\radian}$  & $\SI{0.01}{\metre}$ & $\SI{48}{\metre}$ & $\num{4.40e-04}$ & $\num{1.67e-06}$ & $\num{1.45e-05}$ \\
  $\SI{0.05}{\radian}$  & $\SI{0.01}{\metre}$ & $\SI{72}{\metre}$ & $\num{6.68e-04}$ & $\num{1.76e-06}$ & $\num{1.81e-05}$ \\
  $\SI{0.05}{\radian}$  & $\SI{0.1}{\metre}$  & $\SI{24}{\metre}$ & $\num{9.37e-05}$ & $\num{8.22e-07}$ & $\num{3.21e-06}$ \\
  $\SI{0.05}{\radian}$  & $\SI{0.1}{\metre}$  & $\SI{48}{\metre}$ & $\num{1.14e-04}$ & $\num{8.57e-07}$ & $\num{3.66e-06}$ \\
  $\SI{0.05}{\radian}$  & $\SI{0.1}{\metre}$  & $\SI{72}{\metre}$ & $\num{1.23e-04}$ & $\num{9.52e-07}$ & $\num{4.35e-06}$ \\
  $\SI{0.05}{\radian}$  & $\SI{1}{\metre}$    & $\SI{24}{\metre}$ & $\num{8.32e-05}$ & $\num{8.74e-07}$ & $\num{3.05e-06}$ \\
  $\SI{0.05}{\radian}$  & $\SI{1}{\metre}$    & $\SI{48}{\metre}$ & $\num{7.91e-05}$ & $\num{8.83e-07}$ & $\num{3.05e-06}$ \\
  $\SI{0.05}{\radian}$  & $\SI{1}{\metre}$    & $\SI{72}{\metre}$ & $\num{8.60e-05}$ & $\num{8.88e-07}$ & $\num{3.05e-06}$ \\
  $\SI{0.5}{\radian}$   & $\SI{0.01}{\metre}$ & $\SI{24}{\metre}$ & $\num{4.00e-04}$ & $\num{1.81e-06}$ & $\num{1.50e-05}$ \\
  $\SI{0.5}{\radian}$   & $\SI{0.01}{\metre}$ & $\SI{48}{\metre}$ & $\num{3.90e-04}$ & $\num{1.80e-06}$ & $\num{1.49e-05}$ \\
  $\SI{0.5}{\radian}$   & $\SI{0.01}{\metre}$ & $\SI{72}{\metre}$ & $\num{3.90e-04}$ & $\num{1.80e-06}$ & $\num{1.49e-05}$ \\
  $\SI{0.5}{\radian}$   & $\SI{0.1}{\metre}$  & $\SI{24}{\metre}$ & $\num{3.40e-04}$ & $\num{1.81e-06}$ & $\num{1.18e-05}$ \\
  $\SI{0.5}{\radian}$   & $\SI{0.1}{\metre}$  & $\SI{48}{\metre}$ & $\num{3.69e-04}$ & $\num{1.85e-06}$ & $\num{1.50e-05}$ \\
  $\SI{0.5}{\radian}$   & $\SI{0.1}{\metre}$  & $\SI{72}{\metre}$ & $\num{4.00e-04}$ & $\num{1.82e-06}$ & $\num{1.50e-05}$ \\
  $\SI{0.5}{\radian}$   & $\SI{1}{\metre}$    & $\SI{24}{\metre}$ & $\num{1.03e-04}$ & $\num{1.48e-06}$ & $\num{4.68e-06}$ \\
  $\SI{0.5}{\radian}$   & $\SI{1}{\metre}$    & $\SI{48}{\metre}$ & $\num{1.40e-04}$ & $\num{1.64e-06}$ & $\num{6.23e-06}$ \\
  $\SI{0.5}{\radian}$   & $\SI{1}{\metre}$    & $\SI{72}{\metre}$ & $\num{1.78e-04}$ & $\num{1.71e-06}$ & $\num{7.45e-06}$ \\
\midrule
  \multicolumn{3}{c |}{total} & $\num{6.68e-04}$ & $\num{1.26e-06}$ & $\num{8.93e-06}$ \\
\bottomrule
\end{tabularx}
  \caption{Comparison of the probability distribution of $\pnt{p}_{gt}$ obtained using \ac{MC} sampling and the Gaussian approximation for representative values of the parameters $\sigma_{\si{\metre}}$, $\sigma_{\si{\radian}}$, and $d$.}
  \label{tab:prob_difference}
\end{table}

The maximum, mean and variance of $D_r$ (denoted $\max\left( D_r \right)$, $\bar{D}_p$, and $\sigma_{D_r}$) was calculated over a set of cuboid regions $\mathcal{S} = \left\{ s \right\}$ for each combination of these parameters.
The set $\mathcal{S}$ was chosen to uniformly cover all points $\pnt{p}_{gt}$ sampled for a given combination of parameters by $100 \times 100 \times 100$ equal regions $s \in \mathcal{S}$.
Thus, the empirical probabilities $p_{\text{mc}}\left( \pnt{p}_{gt} \in s \mid \pnt{p}_{m}, \mat{\Sigma}_{\pnt{w}} \right)$ over the set $\mathcal{S}$ can be interpreted as a 3D histogram.
The number of \ac{MC} samples $N=10^5$ and the number of regions $\abs{\set{S}}$ were determined empirically to ensure an accurate representation of the underlying distribution by increasing these metaparameters until the changes were negligible.

The results are summarized in Table~\ref{tab:prob_difference} for a select subset of the considered parameter values.
As evident from the data, the approximation error generally increases with an increase in the orientation uncertainty, which is to be expected as that is the source of non-linearity.
Overall, we conclude that the proposed Gaussian approximation fits the real distribution well for all considered parameter values which represent different \ac{UAV} use-cases.
This approximation is useful for estimating various properties of the problem, as will be discussed in the following sections.

\subsection{Mean position estimation error}
\label{sec:approx_error}

Let us define an error $e_{\text{eucl}}$ of the measured point $\pnt{p}_m$ as its Euclidean distance from the ground-truth point $\pnt{p}_{gt}$ as
\begin{equation}
  e_{\text{eucl}} = \norm{ \pnt{p}_m - \pnt{p}_{gt} } = \norm{ {}_\Delta \pnt{p} },
\end{equation}
where ${}_\Delta \pnt{p} \equiv \pnt{p}_m - \pnt{p}_{gt}$ is the corresponding error vector.
The square of the error can be expressed as
\begin{equation}
  e_{\text{eucl}}^2 = \norm{ {}_\Delta \pnt{p} }^2 = {}_\Delta \pnt{p}\tran {}_\Delta \pnt{p}, \label{eq:eeucl2}
\end{equation}
which is a quadratic form of the multivariate random variable ${}_\Delta \pnt{p}$.
For a general quadratic form $\pnt{\nu}\tran \mat{\Lambda} \pnt{\nu}$ of a random variable $\pnt{\nu}$ with a known expected value $\expected{\pnt{\nu}}$, the expected value is
\begin{equation}
  \expected{\pnt{\nu}\tran \mat{\Lambda} \pnt{\nu}} = \trace{\left( \mat{\Lambda} \mat{\Sigma}_{\pnt{\nu}} \right)} + \pnt{\mu}_{\pnt{\nu}}\tran \mat{\Lambda} \pnt{\mu}_{\pnt{\nu}},
\end{equation}
where $\mat{\Lambda}$ is a general symmetric matrix, $\trace\left( \cdot \right)$ is the trace operator, $\pnt{\mu}_{\pnt{\nu}}$ is the expected value of $\pnt{\nu}$, and $\mat{\Sigma}_{\pnt{\nu}}$ is its covariance matrix~\cite{mathai1992quadratic}.
Furthermore, assuming that $\pnt{\nu}$ is Gaussian, the variance of the quadratic form is
\begin{equation}
  \variance{ \pnt{\nu}\tran \mat{\Lambda} \pnt{\nu} } = 2 \trace{\left( \mat{\Lambda} \mat{\Sigma}_{\pnt{\nu}} \mat{\Lambda} \mat{\Sigma}_{\pnt{\nu}} \right)} + 4 \pnt{\mu}_{\pnt{\nu}}\tran \mat{\Sigma} \mat{\Lambda} \mat{\Sigma} \pnt{\mu}_{\pnt{\nu}}.
\end{equation}
In the case of $e_{\text{eucl}}^2$ as defined in eq.~\eqref{eq:eeucl2}, $\mat{\Lambda}$ is an identity matrix.
Because $\expected{ \pnt{p}_{gt} } = \pnt{p}_m $, the expected value of ${}_\Delta \pnt{p}$ is zero, and thus
\begin{equation}
  \expected{ e_{\text{eucl}}^2 } = \trace\left( \mat{\Sigma} \right).
\end{equation}

Using the Gaussian approximation of the \ac{PDF} of $\pnt{p}_{gt}$ proposed in the previous section, the expected value and variance of $e_{\text{eucl}}^2$ can be estimated as
\begin{align}
  \expected{ e_{\text{eucl}}^2 } &= \trace\left( \mat{\Sigma} \right) \approx \trace\left( \mat{J} \mat{\Sigma}_{\pnt{w}} \mat{J}\tran \right), \\
  \variance{ e_{\text{eucl}}^2 } &= 2 \trace\left( \mat{\Sigma}^2 \right) \approx 2 \trace\left( \left( \mat{J} \mat{\Sigma}_{\pnt{w}} \mat{J}\tran \right)^2 \right).
\end{align}
Furthermore, because $e_{\text{eucl}}^2$ is approximated as a sum of squared random variables with a Gaussian distribution, it follows a Generalized Chi-squared distribution, which is a specialized form of the Gamma distribution.
In general, for a random variable $\rho = \sqrt{\gamma}$ where $\gamma$ is drawn from the Gamma distribution, $\rho$ has the Nakagami-\textit{m} distribution~\cite{souza2007nakagami} described by two parameters $m_{\rho}$ and $\Omega_{\rho}$.
If the expected value and variance of $\gamma$ are known, the two parameters $m_{\rho}$ and $\Omega_{\rho}$ can be obtained as
\begin{align}
  m_{\rho} = \frac{ \expected{\gamma}^2 }{ \variance{\gamma} }, &&
  \Omega_{\rho} = \expected{\gamma}.
\end{align}
Applying this to $e_{\text{eucl}}$, its probability distribution can be estimated (again assuming the Gaussian approximation) as a Nakagami-\textit{m} distribution with
\begin{align}
  m_{e_{\text{eucl}}} &= \frac{ \expected{e_{\text{eucl}}^2}^2 }{ \variance{e_{\text{eucl}}^2} } \approx \frac{ \trace\left( \mat{\Sigma} \right)^2 }{ 2 \trace\left( \mat{\Sigma}^2 \right) }, \\
  \Omega_{e_{\text{eucl}}} &= \expected{e_{\text{eucl}}^2} \approx \trace\left( \mat{\Sigma} \right)^2.
\end{align}
The \ac{PDF}, \ac{CDF}, mean, and variance of the Nakagami-\textit{m} distribution have known analytical forms.
Thus, these approximations can be used to estimate properties of the measured point's error, such as its expected value.

\begin{figure}
  \centering
  \includegraphics[width=0.48\textwidth]{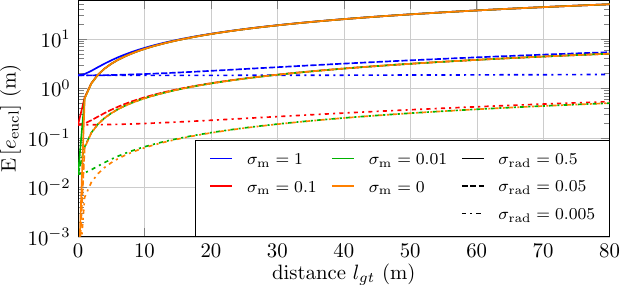}
  \caption{%
    Expected Euclidean error $\expected{e_{\text{eucl}}}$ of a point measured by a \ac{LiDAR} with uncertain pose as a function of distance for different noise parameters~$\sigma_{\si{\metre}}$ and $\sigma_{\si{\radian}}$ obtained using the approximation defined in sec.~\eqref{sec:approx_error}.
  }
  \label{fig:error-dist}
\end{figure}

Values of $\expected{ e_{\text{eucl}} }$ over distance for different distributions of the noise vector $\pnt{w}$ are shown in Fig.~\ref{fig:error-dist}.
It may be observed that for $\sigma_{\text{rad}} \geq \SI{0.05}{\radian}$, the expected error quickly converges to several meters with increasing distance, which shows how crucial accurate orientation measurements are -- especially in outdoor deployments where large distances of the measured objects are common.




\begin{figure}
  \centering
  \includegraphics[width=0.48\textwidth]{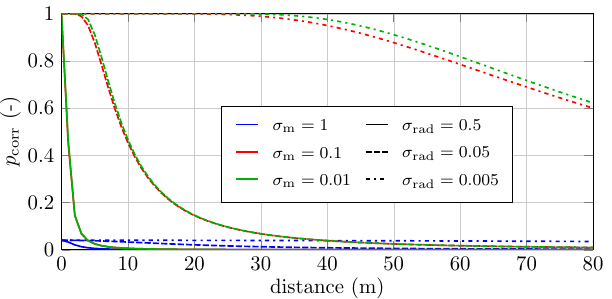}
  \caption{%
    Probability that a point measured by a \ac{LiDAR} with uncertain pose lies in the correct voxel as a function of distance for different noise parameters~$\sigma_{\si{\metre}}$ and $\sigma_{\si{\radian}}$ obtained using the approximation defined in sec.~\ref{sec:approx_corr_prob}.
    The voxel size was set to $\SI{1}{\metre}$.
  }
  \label{fig:probs-dist}
\end{figure}

\subsection{Evaluation of correct voxel update probability}
\label{sec:approx_corr_prob}

In the context of voxel-based mapping with uncertain sensor pose, the probability that the correct voxel is updated by a measured point $\pnt{p}_m$ may be formalized as
\begin{equation}
  p_{\text{corr}} = p\left( v_m = v_{gt} \right) = p\left( \pnt{p}_{gt} \in v_{m} \mid \pnt{p}_m, \mat{\Sigma}_{\text{w}} \right),
\end{equation}
where $v_m = v(\pnt{p}_m)$ is the voxel containing the measured point $\pnt{p}_m$ and $v_{gt} = v(\pnt{p}_{gt})$ is the voxel actually containing the true position of the point $\pnt{p}_{gt}$.
It is assumed that the map is aligned with the world frame $\mathcal{W}$, so that voxels in the map correspond to non-overlapping axis-aligned cubes in $\mathcal{W}$.
The probability $p_{\text{corr}}$ can be evaluated using the approximation from sec.~\ref{sec:meas_cov} as
\begin{equation}
  p_{\text{corr}} = \int_{v_m} f_{\mathcal{N}} \left( \pnt{x}, \pnt{\mu}, \mat{\Sigma} \right) d\pnt{x}, \label{eq:corr_prob}
\end{equation}
where the voxel $v_{m}$ in the integration limits is interpreted as the corresponding axis-aligned cube and $\pnt{f}_{\mathcal{N}} \left( \pnt{x}, \pnt{\mu}, \mat{\Sigma} \right)$ is the multivariate Gaussian \ac{PDF} approximating the distribution of $\pnt{p}_{gt}$, as defined in sec.~\ref{sec:meas_cov}.
This integral does not have an analytical solution, however effective algorithms for its numerical calculation do exist~\cite{genz2004numerical}.

Values of $p_{\text{corr}}$ over distance for different distributions of the noise vector $\pnt{w}$ are shown in Fig.~\ref{fig:probs-dist}.
Similarly as in the previous section, it may be observed that the probability of the correct voxel being updated decreases significantly with $\sigma_{\si{\radian}} \geq \SI{0.05}{\radian}$ at higher distances.
We suspect that this is an often overlooked problem in mapping algorithms when deployed in outdoor environments.
Note that, unlike results from the previous sections, the values in Fig.~\ref{fig:probs-dist} also depend on the alignment of the \ac{LiDAR} ray within the world frame $\mathcal{W}$, represented by $\mat{R}_{gt}$, $\pnt{t}_{gt}$, and $\vec{d}$.
For the sake of simplicity, we have assumed $\mat{R}_{gt} = \mat{I}$, $\pnt{t}_{gt} = \pnt{0}$, and $\vec{d} = \bemat{1 & 0 & 0}\tran$.

\begin{figure}
  \centering
  \includegraphics[width=0.48\textwidth]{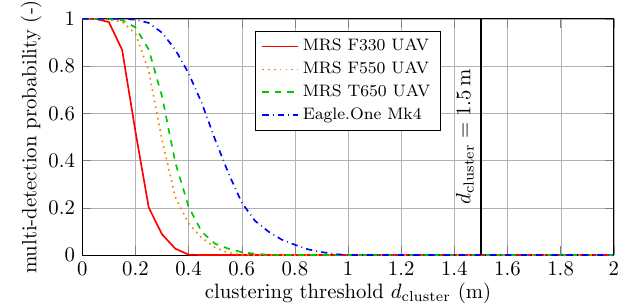}
  \caption{%
    Probability of detecting a single \ac{UAV} as multiple targets (if more than one point of the target was sampled) for different clustering thresholds obtained using \ac{MC} simulations.
    The Ouster OS1-128 \ac{LiDAR} and four different \ac{UAV} models were considered.
    Size of the targets ranges from \SI{0.33}{\metre} to \SI{1.6}{\metre} approx. diameter (see~\cite{HertJINTHW_paper} for more details).
    Value of the threshold chosen for the experiments is marked with a black vertical line.
  }
  \label{fig:graph_multidets}
\end{figure}

\rone{%
\subsection{Clustering tolerance}
\label{sec:cluster_misclass}
Because both the detection and tracking algorithms operate on point clusters, they are sensitive to the choice of the clustering threshold $d_{\text{cluster}}$.
If $d_{\text{cluster}}$ is smaller than sparsity of the sampled points corresponding to an object, it will be separated into multiple clusters, resulting in multiple detections.
On the other hand, if $d_{\text{cluster}}$ is larger than the closest distance between two objects, they will be clustered and detected as a single object.

More formally, let us define the sparsity $s\left(\set{C}\right)$ of a cluster $\set{C}$ as
\begin{equation}
  s\left(\set{C}\right) = \max_{\pnt{p}_i \in\set{C}} \min_{\pnt{p}_j \in\set{C} \setminus \pnt{p}_i} \norm{ \pnt{p}_i - \pnt{p}_j },
\end{equation}
which is the maximal distance between two nearest neighbors in the cluster, and let $\set{C}_A^*$ be a set of points from a single scan corresponding to object A and $\set{C}_B^*$ to object B.
A single object is then detected as multiple if
\begin{equation}
  d_{\text{cluster}} < s\left(\set{C}_A^*\right),
\end{equation}
and two objects will result in a single detection if
\begin{equation}
  d_{\text{cluster}} > \min_{\left(\pnt{p}_i, \pnt{p}_j\right) \in \set{C}_A^* \times \set{C}_B^*} \norm{\pnt{p}_i - \pnt{p}_j}.
\end{equation}

In the case of \ac{UAV} detection, sparsity of the sampled points depends on the shape of the target, geometrical distribution of rays of the \ac{LiDAR} sensor, and their relative pose.
If the target's shape and the sensor's ray distribution are known, distribution of the sparsity of sampled points can be obtained using \ac{MC} simulation over varying relative poses.
This way, the probability of detecting a single \ac{UAV} as multiple targets for different $d_{\text{cluster}}$ can be estimated, which provides a lower bound for its selection.
The upper bound of $d_{\text{cluster}}$ is given by the minimal assumed distance between trajectories of two targets or of a target and background obstacles.

To select the clustering threshold for the considered range of targets and the used \ac{LiDAR} sensor, the multi-detection probability was evaluated for four representative \ac{UAV} platforms with dimensions ranging from \SI{0.33}{\metre} to \SI{1.6}{\metre} approx. diameter (see Fig.~\ref{fig:graph_multidets}).
Based on the results and assuming a minimal distance of general \acp{UAV} from obstacles for a safe outdoor flight of \SI{1.5}{\metre}, we recommend $d_{\text{cluster}} = \SI{1.5}{\metre}$.

}

\section{Experimental Evaluation}
\label{sec:experiments}

\begin{figure}[t]
  \centering
  \includegraphics[width=0.48\textwidth]{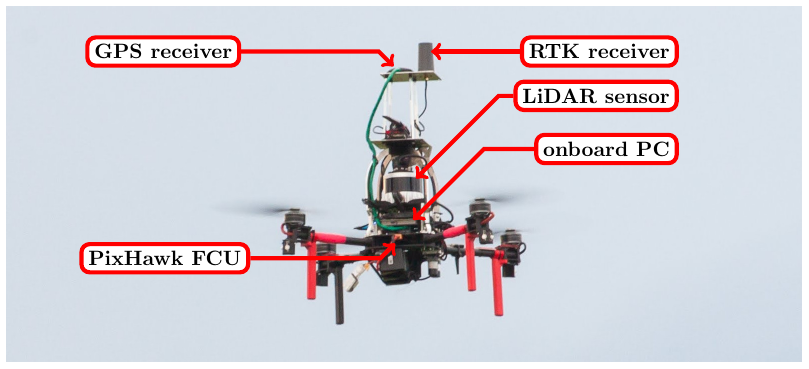}
  \caption{%
    Schematic of the \ac{UAV} platform used in the experiments.
  }
  \label{fig:tarot}
\end{figure}

Several real-world and simulated experiments, aimed to evaluate performance of the presented algorithms under realistic conditions, are presented in this section.
Parameter values of the algorithms during the experiments are listed in Table~\ref{tab:exp_params}.
Unless otherwise specified, the experiments were conducted using the MRS UAV platform based on the Tarot T650 frame (with an approximate largest width between rotors of $\SI{0.65}{\metre}$) equipped with a PixHawk flight control unit including an \ac{IMU} and an Intel NUC computer~\cite{HertJINTHW_paper} (see Fig.~\ref{fig:tarot}).
The onboard computer runs the MRS UAV system~\cite{baca2021mrs} for self-localization, stabilization, and trajectory tracking.
The parameters of the \acp{LiDAR} used in the simulations are summarized as stated by the manufacturer in Table~\ref{tab:os_params}.
Videos from all experiments are available online\footnote{\href{https://mrs.felk.cvut.cz/flying-object-detection}{\url{https://mrs.felk.cvut.cz/flying-object-detection}}}.


\begin{table}[t]
  \footnotesize
  \begin{tabularx}{0.48\textwidth}{l l l l l l l l}
    \toprule
    \multicolumn{7}{c}{\textbf{Occupancy mapping}} \\
    \midrule
    $s_v$             & $g_{\text{occ}}$  & $g_{\text{unk}}$  & $g_{\text{free}}$ & $G_{\text{conf}}$ & $G_{\text{tent}}$ & $G_{\text{unc}}$ \\
    \SI{0.25}{\metre} & 0                 & -740              & -1000             & -0.1              & -300              & -750             \\
    \toprule
    \multicolumn{7}{c}{\textbf{Clustering, raycasting, and separate voxel removal}} \\
    \midrule
    $d_{\text{cluster}}$    & $d_{\text{close}}$  & $d_{\text{search}}$ & $d_{\text{max}}$  & $n_{\text{conf}, \text{min}}$   & $w_{\text{int}}$ \\
    \rone{\SI{1.5}{\metre}} & \rone{\SI{1.5}{\metre}}    & \SI{3}{\metre}      & \SI{20}{\metre}   & 24                              & $0.003$\\
  \end{tabularx}
  \begin{tabularx}{0.48\textwidth}{l l l l l l l l}
    \toprule
    \multicolumn{6}{c}{\textbf{Multi-target tracking}} \\
    \midrule
    $N_{\text{buf}}$  & $c_r$ & $r_{\text{min}}$  & $r_{\text{max}}$  & $d_{\text{min}}$ & Z \\
    10                & 1.5   & \SI{2.5}{\metre}  & \SI{5}{\metre}    & \SI{1}{\metre}   & \SI{0.3}{\metre} \\
    \midrule
    $\Xi_{\pnt{r}}$   & $\Xi_{\pnt{v}}$             & $\Xi_{\pnt{a}}$                     & $P_{0,\pnt{r}}$   & $P_{0,\pnt{v}}$ & $P_{0,\pnt{a}}$ \\
    \SI{0.01}{\metre} & \SI{0.2}{\metre\per\second} & \SI{0.3}{\metre\per\second\squared} & \SI{0.3}{\metre}  & \SI{1}{\metre\per\second} & \SI{1}{\metre\per\second\squared} \\
    %
    \bottomrule
  \end{tabularx}
  \caption{Parameter values of the presented algorithms used in the experiments.}
  \label{tab:exp_params}
\end{table}



\begin{table}[t]
  \footnotesize
  \begin{tabularx}{0.48\textwidth}{l X X X X X l}
    \toprule
    \textbf{model} & \textbf{horiz. rays} & \textbf{vertical rays} & \textbf{vertical FOV} & \textbf{scan rate} & \textbf{max. range} & \textbf{precision}\\
    \midrule
    OS1-128 Rev. C & 1024 & 128 & $\ang{\pm 22.5}$ & \SI{10}{\hertz} & \SI{100}{\metre} & $\SI{\pm 5}{\centi \meter}$ \\ 
    OS0-128 Rev. C & 1024 & 128 & $\ang{\pm 45}$ & \SI{10}{\hertz} & \SI{45}{\metre} & $\SI{\pm 5}{\centi \meter}$ \\ 
    \bottomrule
  \end{tabularx}
  \caption{Parameters of the sensors used in the experiments according to the manufacturer.}
  \label{tab:os_params}
\end{table}



\subsection{Simulated experiments}
\label{sec:simulations}

To establish a baseline of the algorithms' performance under ideal conditions, a set of simulations was performed with no noise burdening the observer's self-localization and the \ac{LiDAR} sensing.
A second set of simulations with noisy measurements was executed where parameters of the noise were selected according to the stated characteristics of the sensors employed in the real-world experiments described in the following section.
Using notation from the previous sections, the corresponding covariance matrix of the noise was
\begin{equation}
  \begin{split}
    \mat{\Sigma}_{\pnt{w}} = \diag\left( \right. &\SI{0.03}{\metre}, 
    ~\SI{0.05}{\metre}, ~\SI{0.05}{\metre}, ~\SI{0.05}{\metre}, \\
    &\left. \SI{0.005}{\radian}, ~\SI{0.005}{\radian}, ~\SI{0.005}{\radian} \right)^2.
  \end{split}
\end{equation}

\begin{figure}
  \centering
  \includegraphics[width=0.48\textwidth]{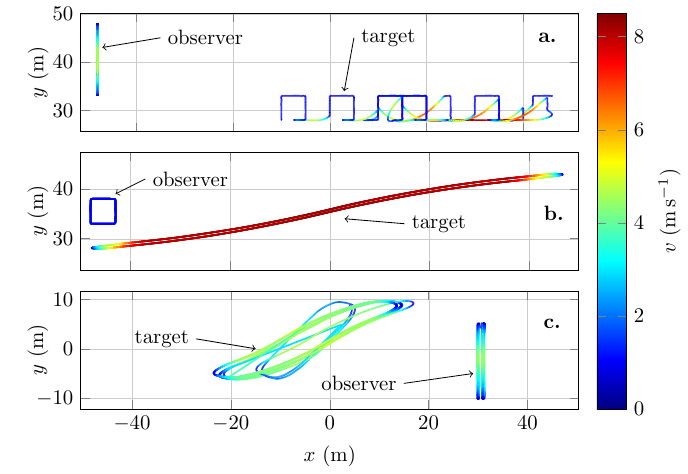}
  \caption{
    Top-down view of example trajectories of the observer and target \acp{UAV} used in the simulated (\textbf{a.}, \textbf{b.}) and real-world experiments (\textbf{c.}).
    Color of the lines denotes velocity.
    Altitude of the trajectories varied between \SI{4}{\metre} and~\SI{10}{\metre}.
  }
  \label{fig:exp_trajectories}
\end{figure}

The observer was equipped with a simulated \ac{LiDAR} of the same ray layout and density as the Ouster OS1-128 sensor.
Both the target and the observer followed various trajectories with changing mutual distance, velocities, and accelerations, as illustrated in Fig.~\ref{fig:exp_trajectories} and summarized in Table~\ref{tab:exp_trajparams}.
The simulations were executed in the Gazebo simulator\footnote{\href{http://gazebosim.org}{\url{http://gazebosim.org}}}.
The environment where the real-world experiments took place was mapped using a 3D scanner and imported into the simulator to ensure high fidelity to real-world conditions.

Results of the experiments were evaluated using the following metrics:
\begin{align}
  T^* &= \argmin_{T \in \mathcal{T}} \norm{ \hat{\pnt{r}}_T - \pnt{r} }, \\
  e_{\pnt{r}} &= \norm{ \hat{\pnt{r}}_{T^*} - \pnt{r} },   \\
  e_{\vec{v},\text{mag}} &= \abs{ \norm{ \hat{\vec{v}}_{T^*} } - \norm{ \vec{v} } },   \\
  e_{\vec{v},\text{ang}} &= \arccos{\left( \frac{ \hat{\vec{v}}_{T^*} \cdot \vec{v} }{ \norm{ \hat{\vec{v}}_{T^*} } \cdot \norm{ \vec{v} } } \right)},   \\
  \text{accuracy} &= \frac{ \text{TP} + \text{TN} }{ \text{TP} + \text{TN} + \text{FP} + \text{FN} }, \\
  \text{recall} = \frac{ \text{TP} }{ \text{TP} + \text{FN} },
  & \hspace{3em} \text{precision} = \frac{ \text{TP} }{ \text{TP} + \text{FP} },
\end{align}
where $T^*$ is the closest track, $e_{\pnt{r}}$ is position error, $e_{\vec{v},\text{mag}}$ is velocity magnitude error, $e_{\vec{v},\text{ang}}$ is velocity angle error, $\text{TP}$ are true positives, $\text{TN}$ are true negatives, $\text{FP}$ are false positives, and $\text{FN}$ are false negatives.
A detection is considered a true positive if the distance of its estimated position from the ground-truth is less than \SI{3}{\metre}.
The position error of the raw detections was evaluated analogously to the tracks.
The results are presented in the graphs in Fig.~\ref{fig:exp_results} and in Table~\ref{tab:exp_results}.
Furthermore, timings of the different modules were measured during the simulation and are presented in Table~\ref{tab:exp_timings}.
\rone{The measured timings correspond to an average output rate of \SI{8.9}{\hertz} for the detection and \SI{49.8}{\hertz} for the multi-target tracking with the used processor.}

It may be observed that even though all sensors in the noiseless experiment were ideal, there is a non-zero error in the detected position of the target.
This error is caused by a biased sampling of the target's shape by the observer's \ac{LiDAR} sensor, which is always observed only from one side during a single detection.
Therefore, points from the observed side of the target are overrepresented in a single \ac{LiDAR} scan and the resulting position estimate (obtained as their mean) is biased.
The value of this bias depends on the target's shape, but it is always in the range $\interval{0}{a}$, where $a$ is a radius of its circumscribed sphere.
This corresponds well with the dimensions of the target used in the experiments, for which $a = \SI{0.325}{\metre}$.
Results of the simulation with noise show an increase in the error metrics in comparison to the noiseless results.
According to the results in sec.~\ref{sec:approx_error}, the mean position error is higher than the theoretical expected error, although the difference is not significant, which we attribute to the same sampling bias effect as for the noiseless simulations.
Finally, it is worth noting that we have observed no clear correlation between the velocity or acceleration of the \acp{UAV} and error of the position estimation.



\begin{table}
  \footnotesize
  \begin{tabularx}{0.48\textwidth}{l | X X X | X}
    \toprule
    \textbf{parameter}          & \multicolumn{3}{ c |}{\textbf{simulated}}                  & \multirow{2}{\hsize}{\textbf{real-world}} \\
                                & \textbf{no noise} & \textbf{with noise} & \textbf{multi-UAV} &                     \\
    \midrule
    max. mutual distance        & \SI{90}{\metre}                     & \SI{90}{\metre}                     & \SI{16}{\metre}                     & \SI{55}{\metre} \\
    max. velocity (target)      & \SI{8.2}{\metre\per\second}         & \SI{8.2}{\metre\per\second}         & \SI{1}{\metre\per\second}         & \SI{4.8}{\metre\per\second} \\
    max. velocity (observer)    & \SI{5.5}{\metre\per\second}         & \SI{5.5}{\metre\per\second}         & \SI{1}{\metre\per\second}         & \SI{4.5}{\metre\per\second} \\
    max. acceleration           & \SI{4.5}{\metre\per\second\squared} & \SI{4.5}{\metre\per\second\squared} & \SI{2.5}{\metre\per\second\squared} & \SI{3.0}{\metre\per\second\squared} \\
    total length (target)       & \SI{1331}{\metre}                   & \SI{1782}{\metre}                   & \SI{367}{\metre}                   & \SI{1060}{\metre} \\
    total length (observer)     & \SI{755}{\metre}                    & \SI{880}{\metre}                    & \SI{276}{\metre}                   & \SI{810}{\metre} \\
    total duration              & \SI{355}{\second}                   & \SI{400}{\second}                   & \SI{517}{\second}                   & \SI{375}{\second} \\
    \bottomrule
  \end{tabularx}
  \caption{Characterization of the trajectories used in the experiments.}
  \label{tab:exp_trajparams}
\end{table}



\begin{table}
  \footnotesize
  \begin{tabularx}{0.48\textwidth}{l | X | X l}
    \toprule
                        & \multirow{2}{\hsize}{\textbf{position error} (\si{\metre})}  & \multicolumn{2}{c}{\textbf{velocity error}} \\
                        &   & \textbf{magnitude} (\si{\metre\per\second}) & \textbf{angle} (\si{\radian}) \\
    \midrule
    \multicolumn{4}{c}{\textbf{Simulated experiments (no noise)}} \\
    \midrule
    \textbf{detector} & 0.26 (0.14) & N/A & N/A \\
    \textbf{tracker} & 0.28 (0.21) & 0.68 (0.82) & 0.31 (0.54) \\
    \midrule
    \multicolumn{4}{c}{\textbf{Simulated experiments (with noise)}} \\
    \midrule
    \textbf{detector} & 0.43 (0.24) & N/A & N/A \\
    \textbf{tracker} & 0.36 (0.19) & 0.71 (0.68) & 0.21 (0.48) \\
    \midrule
    \multicolumn{4}{c}{\textbf{Simulated experiments (multi-UAV)}} \\
    \midrule
    \textbf{detector} & 0.11 (0.07) & N/A & N/A \\
    \textbf{tracker} & 0.13 (0.12) & 0.11 (0.10) & 0.30 (0.44) \\
    \midrule
    \multicolumn{4}{c}{\textbf{Real-world experiments}} \\
    \midrule
    \textbf{detector} & 0.25 (0.11) & N/A & N/A \\
    \textbf{tracker} & 0.29 (0.15) & 0.61 (0.64) & 0.18 (0.13) \\
    \bottomrule
  \end{tabularx}
  \caption{%
    Means and standard deviations (in parentheses) of the position and velocity estimation errors during the experiments.
    The multi-UAV and real-world results have a lower error than the simulated experiments with noise due to an unbalanced data set caused by a limited detection distance (and smaller targets in the multi-UAV case).
    For the multi-UAV experiments, the velocity angle error is only calculated for the moving target to avoid singularities with near zero-length vectors, the other metrics are averaged over all targets.
    }
  \label{tab:exp_results}
\end{table}



\begin{table}
  \footnotesize
  \begin{tabularx}{0.48\textwidth}{l | X X X X}
    \toprule
    \textbf{algorithm / routine}  & \textbf{mean}  & \textbf{standard dev.}  & \textbf{max. value}  & \vspace{-0.85em}\rone{\textbf{max. rate}} \\
    \midrule
                        \multicolumn{4}{c}{\textbf{mapping \& detection}} & \vspace{-0.85em}\rone{\SI{8.9}{\hertz}} \\
    \midrule
    clustering \& classification     & \SI{112.3}{\milli\second}     & \SI{52.1}{\milli\second}      & \SI{304.0}{\milli\second} \\
    raycasting      & \SI{319.5}{\milli\second}     & \SI{57.2}{\milli\second}      & \SI{532.0}{\milli\second} \\
    separate bg. clusters removal   & \SI{159.6}{\milli\second}     & \SI{43.3}{\milli\second}      & \SI{312.0}{\milli\second} \\
    \midrule
                        \multicolumn{4}{c}{\textbf{multi-target tracking}} & \vspace{-0.85em}\rone{\SI{49.8}{\hertz}} \\
    \midrule
    \texttt{newPointCloud()}  & \SI{20.1}{\milli\second}      & \SI{12.7}{\milli\second}      & \SI{76.0}{\milli\second} \\
    \texttt{newDetections()}  & \SI{6.4}{\milli\second}       & \SI{3.4}{\milli\second}       & \SI{16.0}{\milli\second} \\
    \texttt{updateTrack()}    & \SI{2.7}{\milli\second}       & \SI{2.0}{\milli\second}       & \SI{12.0}{\milli\second} \\
    \bottomrule
  \end{tabularx}
  \caption{%
    Execution time of different elements of the proposed system with input data from a simulated Ouster OS1-128 ($1024\times 128$ rays at $\SI{10}{\hertz}$) measured on an AMD Ryzen 3900X 12-core CPU with \SI{32}{\gibi \byte} RAM.
    }
  \label{tab:exp_timings}
\end{table}



\subsection{Multi-UAV experiments}
\label{sec:multiuav}


\begin{figure}
  \centering
  \begin{subfigure}[t]{0.48\textwidth}
    \centering
    \includegraphics[width=\textwidth]{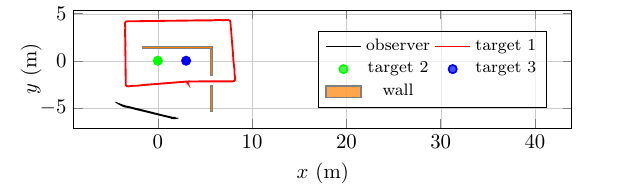}
    \caption{%
      A top-down view.
    }
  \end{subfigure}%
  \vspace{1em}

  \begin{subfigure}[t]{0.48\textwidth}
    \centering

    \begin{tikzpicture}
      \node[anchor=north west,inner sep=0] (a) at (0, 0)
      {
        \includegraphics[width=\textwidth, trim=0cm 2cm 0cm 3cm, clip=true]{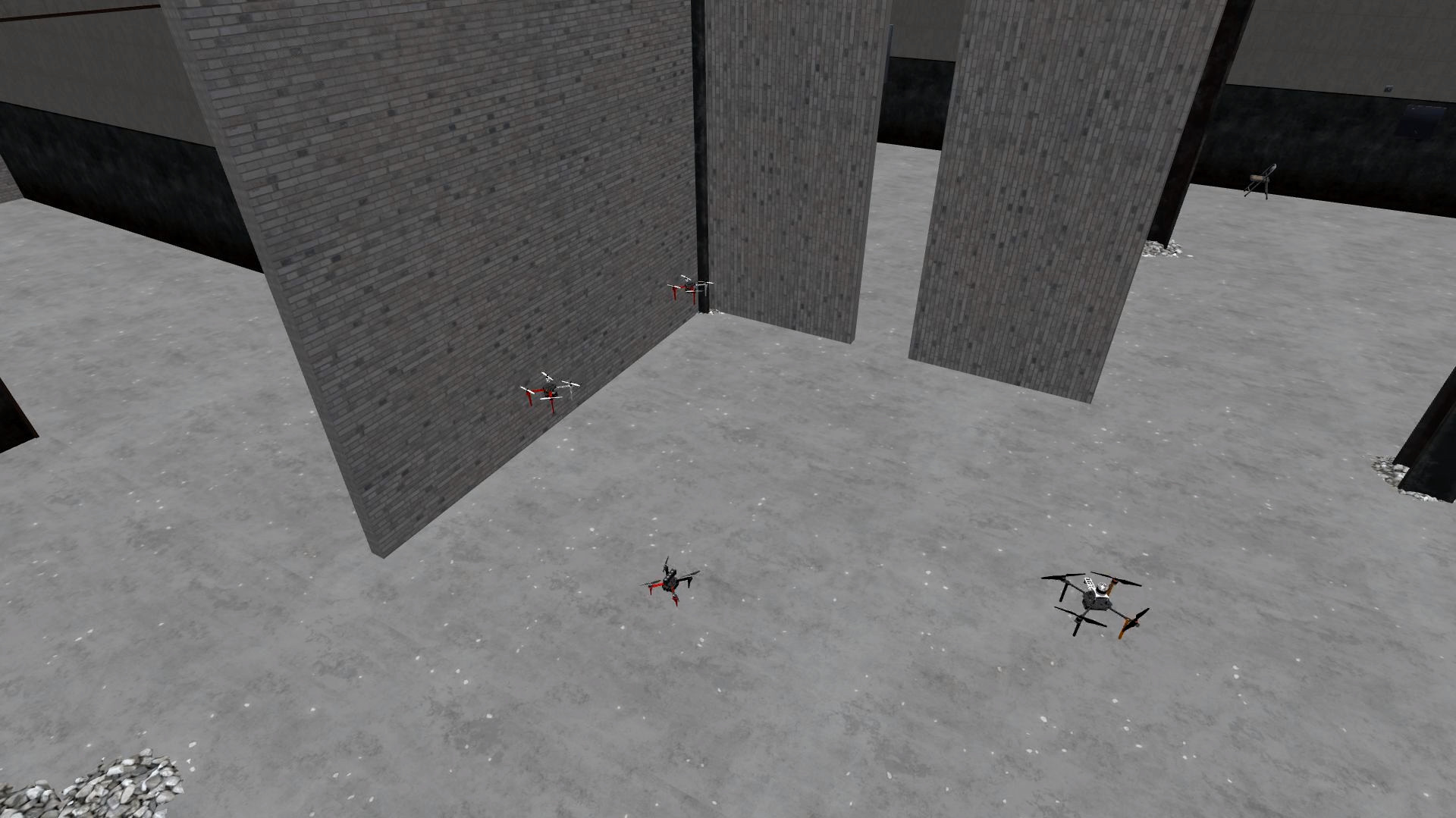}
      };

    \node[line width=2pt, draw, red, text=black, rounded corners, fill=white, anchor=west] (observer) at (6.7, -2.2) {\textbf{observer}};
      \draw[red, line width=2pt, ->] (observer.south) -- ++(-0.5, -0.5);

    \node[line width=2pt, draw, red, text=black, rounded corners, fill=white, anchor=west] (target1) at (1.8, -3.5) {\textbf{target 1}};
      \draw[red, line width=2pt, ->] (target1.east) -- ++(0.5, 0.3);

    \node[line width=2pt, draw, red, text=black, rounded corners, fill=white, anchor=west] (target2) at (1.0, -2.0) {\textbf{target 2}};
      \draw[red, line width=2pt, ->] (target2.east) -- ++(0.5, 0.0);

    \node[line width=2pt, draw, red, text=black, rounded corners, fill=white, anchor=west] (target3) at (1.95, -0.95) {\textbf{target 3}};
      \draw[red, line width=2pt, ->] (target3.east) -- ++(0.5, -0.3);
    \end{tikzpicture}

    \caption{%
      A perspective view.
    }
  \end{subfigure}%
  \caption{
    Illustration of the multi-UAV simulated experiment setup with highlighted trajectories of the \acp{UAV}.
    Trajectory of one of the targets is partially obstructed from the observer's point of view to test track initialization and loss.
  }
  \label{fig:multiuav_photos}
\end{figure}
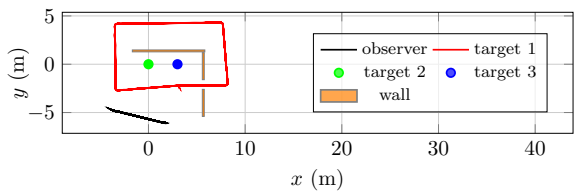
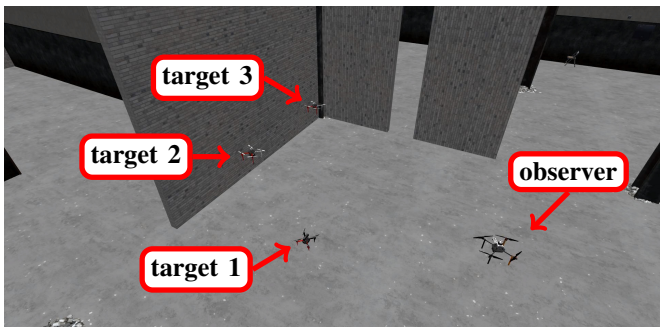



\begin{figure}
  \centering
  \includegraphics[width=0.48\textwidth]{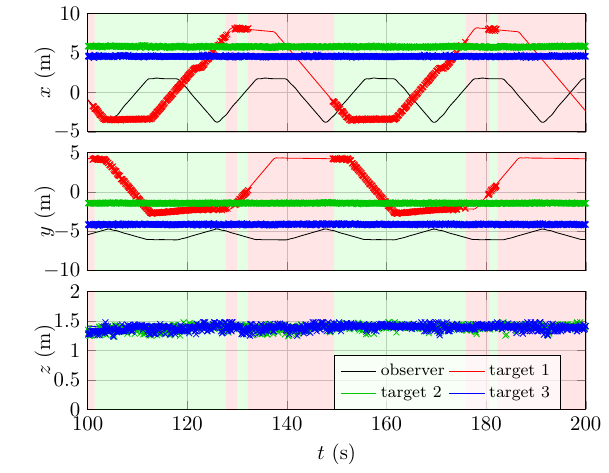}
  \caption{
    Per-axis graphs of the trajectories of the \acp{UAV} (solid lines) and the corresponding tracks (marked with crosses) in the multi-UAV simulated experiment.
    Background color of the graphs marks times when target 1 was within line of sight (green) or hidden behind the wall (red).
    Only a \SI{100}{\second} window is shown for conciseness, as the trajectories were periodical and the results were similar during the whole experiment.
  }
  \label{fig:multiuav_graphs}
\end{figure}


The detection and tracking system presented in this paper is capable of tracking multiple targets simultaneously without hampering its performance, is robust to variations in environments, and can work with different sensor configurations.
To demonstrate this, a simulated experiment with one observer and three target \acp{UAV} in an indoor area was devised.
In this case, the observer was equipped with a simulated \ac{LiDAR} having the same parameters as the Ouster OS0-128 sensor (see Table~\ref{tab:os_params}).
For self-localization, it used the LOAM-SLAM \cite{zhang2014loam}, which closely emulates conditions of a real-world indoor deployment.
\rone{To accommodate the different conditions, $d_{\text{cluster}} = \SI{0.3}{\metre}$ and $d_{\text{close}} = \SI{0.3}{\metre}$ was used as opposed to the other experiments, which were performed outdoors.}

Two of the targets were stationary (based on the DJI F330 platform) and one was moving (using the DJI F450 platform).
To further show the robustness of the track association algorithm, part of the trajectory of the moving target led behind a solid wall so that the corresponding track was lost and recovered during parts of the trajectory.
This is visualized in Fig.~\ref{fig:multiuav_photos}.
As evident in Fig.~\ref{fig:multiuav_graphs}, the system successfully tracked all the targets, including the reappearing track, without any cross-association.
Results of the simulation are summarized in Table~\ref{tab:exp_results}.



\subsection{Real-world experiments}

\begin{figure}
  \centering
  \begin{subfigure}[t]{0.48\textwidth}
    \centering
    \includegraphics[width=\textwidth]{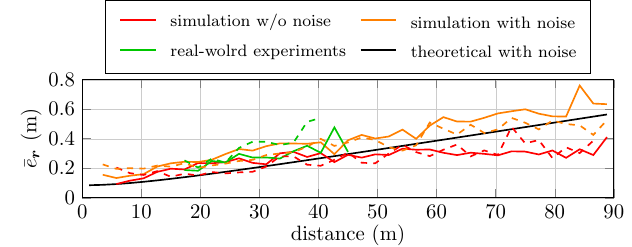}
    \caption{%
    Mean position error over distance.
    }
    \label{fig:results_err_over_dist}
  \end{subfigure}%

  \begin{subfigure}[t]{0.48\textwidth}
    \centering
    \includegraphics[width=\textwidth]{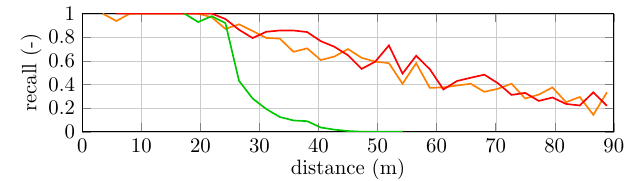}
    \caption{%
    Detection recall over distance.
    }
  \end{subfigure}%
  \caption{Results from the simulated and real-world experiments.
  In (\subref{fig:results_err_over_dist}), the dashed lines correspond to output of the multi-target tracker and the black line shows the expected error caused by the noise in the sensor's pose measurements (corresponding to the simulations with noise).
  Multi-UAV experiments are omitted due to the significantly different distance scale.
  }
  \label{fig:exp_results}
\end{figure}

Several outdoor experiments with one target \ac{UAV} and one observer \ac{UAV} were executed to evaluate the detection system under realistic conditions close to the intended deployment.
Both the target and the observer were equipped with RTK-GPS, which was used as a source of ground truth localization, and the observer carried the Ouster OS1-128 sensor.
Similarly as in the simulated experiments, the \acp{UAV} were following agile pre-planned trajectories (see Fig.~\ref{fig:exp_trajectories} and Table~\ref{tab:exp_trajparams}), although the dynamic constraints were reduced and the minimal distance was limited to \SI{15}{\metre} for safety reasons.
The same metrics as in the simulated experiments were measured and the results are presented in Fig.~\ref{fig:exp_results} and Table~\ref{tab:exp_results}.

It may be observed that the recall is overall worse than in the simulations.
We assume that this is because of a lower reflectivity of the real target (which is modeled as ideal in the simulations), causing the ray reflections to not be registered by the sensor at higher distances.
Unlike in the simulations, the probability distribution of $\pnt{w}$ changes based on multiple factors, such as motion of the observer, distance of the target, or fluctuations in the magnetic field measurements of the onboard magnetometer.
This complicates evaluation of the expected error of the detected position over distance.
However, the observed data overall corresponds well with the simulations where noise was emulated based on the sensor parameters reported by the manufacturers, as indicated by the graphs in Fig.~\ref{fig:exp_results}.



\rone{%
\subsection{Deep learning approaches}
\label{sec:deep-learning}
For completeness, we provide a discussion and a short evaluation of deep learning-based methods applicable to the detection of flying objects using \ac{LiDAR} data.
To the best of our knowledge, there are no works directly tackling this problem using deep learning, but there are several state-of-the-art neural networks designed for general point cloud-based classification, detection, semantic segmentation, and processing as discussed in sec.~\ref{sec:sota}.
We argue that such methods are not suitable for \ac{UAV} detection beyond a relatively short range, because at longer distances the target is sampled only by a low number of rays and the sampled points do not provide sufficient spatial features that could be exploited by a neural network to distinguish a \ac{UAV} from a cluster of several solitary points corresponding to the background (see Fig.~\ref{fig:detection_points}).

To test this hypothesis, we chose the PointNet++ as a representative deep learning approach for point cloud-based 3D feature extraction~\cite{qi2017PointNet2}.
A dataset of 5455 full scans using two \ac{LiDAR} types (listed in Table~\ref{tab:os_params}), three different \ac{UAV} targets, and in two different outdoor environments (semi-urban and a field) was created.
Each point in the dataset was labeled as either a \ac{UAV} or background, and the data was split to training (4242 scans) and testing (1213 scans) sets.
For completeness, we make the dataset publicly available online\footnote{\href{https://mrs.felk.cvut.cz/flying-object-detection}{\url{https://mrs.felk.cvut.cz/flying-object-detection}}}.
Using this dataset, the semantic segmentation variant of the PointNet++ neural network was trained for 200 epochs and evaluated using the best obtained weights.
The results are presented in Table~\ref{tab:pointnet_results}.


\begin{table}
\footnotesize
  \begin{tabularx}{\linewidth}{X | X X | X X}
\toprule
    \multicolumn{1}{r |}{\textbf{class}}        & \multicolumn{2}{c |}{\textbf{\ac{UAV}}}                       & \multicolumn{2}{c}{\textbf{background}}  \\
    \textbf{dataset}                            & \textbf{precision}                      & \textbf{recall}     & \textbf{precision}    & \textbf{recall}     \\
\midrule
    \multicolumn{5}{c}{\textbf{PointNet++~\cite{qi2017PointNet2}}} \\
\midrule
    \textbf{training}                           & $\phantom{0}\SI{3.9}{\percent}$         & \SI{92.6}{\percent} & \SI{>99.9}{\percent}  & \SI{99.6}{\percent} \\
    \textbf{testing}                            & $\phantom{0}\SI{4.0}{\percent}$         & \SI{48.8}{\percent} & \SI{>99.9}{\percent}  & \SI{99.7}{\percent} \\
\midrule
    \multicolumn{5}{c}{\textbf{Ours}} \\
\midrule
    \textbf{training}                           & \SI{96.7}{\percent}         & \SI{97.3}{\percent} & \SI{>99.9}{\percent}  & \SI{>99.9}{\percent} \\
    \textbf{testing}                            & \SI{92.8}{\percent}         & \SI{61.5}{\percent} & \SI{>99.9}{\percent}  & \SI{>99.9}{\percent} \\
\bottomrule
\end{tabularx}
  \caption{Per-point classification results of the proposed detector and of the PointNet++ neural network~\cite{qi2017PointNet2} on full \ac{LiDAR} scans from our UAV detection dataset.
  For a clear comparison, we report the performance of our detector on the same data splits (training and testing) even though it does not rely on training.
  Note that the very high recall and precision of the \textbf{background} class are due to strong class imbalance in the data as most points in the scans belong to the background\rtwo{, which is an inherent property of the problem.}}
  \label{tab:pointnet_results}
\end{table}



\begin{figure}
  \centering
  \begin{subfigure}[t]{0.48\textwidth}
    \centering
    \includegraphics[width=\textwidth]{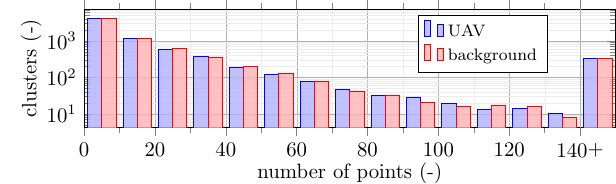}
    \caption{%
      Distribution of the number of points within a cluster in the dataset.
    }
    \label{fig:pointnet_cls_dataset}
  \end{subfigure}%

  \begin{subfigure}[t]{0.48\textwidth}
    \centering
    \includegraphics[width=\textwidth]{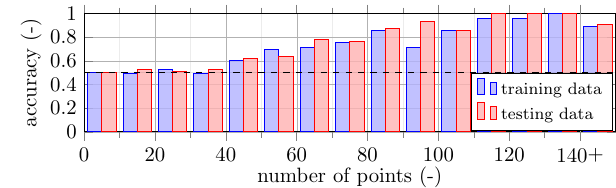}
    \caption{%
      Classification accuracy over the number of points in the cluster. The dashed vertical line denotes the \SI{50}{\percent} accuracy threshold corresponding to a random classifier.
    }
    \label{fig:pointnet_cls_acc}
  \end{subfigure}%
  \caption{%
    Results of the PointNet++ neural network trained for classification of point clusters on our \ac{UAV} cluster classification dataset.
    Because the results for clusters with a high number of points are similar, the last bin in both histograms represents all clusters with more than 140 points for conciseness.
  }
  \label{fig:pointnet_cls}
\end{figure}


It may be observed that although the recall of PointNet++ for the \ac{UAV} class is relatively high, the precision is low, which is due to a high number of false positives.
After analyzing the data, we conclude that the neural network learned to detect small clusters of points that are surrounded by free space.
Therefore, the target \ac{UAV} is usually successfully classified, but also many points corresponding to the background are incorrectly classified as a \ac{UAV}, far outnumbering the \ac{UAV} points.
This supports our original hypothesis.
We have also tried pre-training the network using the ScanNet dataset~\cite{dai2017ScanNetRichlyAnnotated3D} and then fine-tuning using our \ac{UAV} detection dataset, changing the number of input points of the network, and other minor modifications of the PointNet++ architecture and training procedure, but always obtained similar results, which we attribute to the fundamental problem mentioned above.

\rtwo{%
To further evaluate the capability of the neural network to discriminate objects given a low number of points, we created a \ac{UAV} classification dataset sampled from real-world data similarly as the \ac{UAV} detection dataset.
The dataset comprises of 13728 clusters (split to 10982 training and 2746 testing clusters) with each cluster containing points corresponding either to a \ac{UAV} or to background.
The dataset is balanced so that the distribution of the number of points in the clusters is matched for both classes (see Fig.~\ref{fig:pointnet_cls_dataset}).
The classification variant of PointNet++~\cite{qi2017PointNet2} was trained on this data and the results of the best obtained weights are presented in Fig.~\ref{fig:pointnet_cls_acc}.
The classification accuracy is close to \SI{50}{\percent} for clusters with less than 40 points, so the neural network behaves as a random classifier.
The accuracy only starts rising for clusters with more than 40 points, indicating that PointNet++ requires at least approximately 40 unique input points to obtain meaningful results, which corresponds to a target closer than approximately \SI{6}{\metre} with our configuration of sampling density and target size.
}


\begin{figure*}
  \newcommand{\detPtsSz}{0.18\textwidth}
  \centering
  \begin{subfigure}[t]{\detPtsSz}
    \centering
    \includegraphics[width=\textwidth]{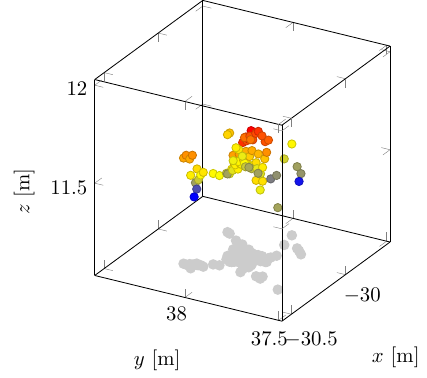}
    \caption{%
      UAV, $l=\SI{4}{\metre}$, 79 points.%
    }
  \end{subfigure}%
  ~
  \begin{subfigure}[t]{\detPtsSz}
    \centering
    \includegraphics[width=\textwidth]{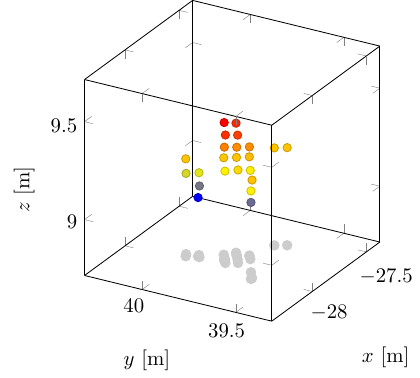}
    \caption{%
      UAV, $l = \SI{10}{\metre}$, 23 points.%
    }
  \end{subfigure}%
  ~
  \begin{subfigure}[t]{\detPtsSz}
    \centering
    \includegraphics[width=\textwidth]{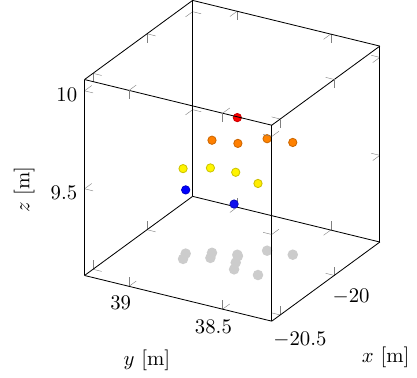}
    \caption{%
      UAV, $l = \SI{20}{\metre}$, 11 points.%
    }
  \end{subfigure}%
  ~
  \begin{subfigure}[t]{\detPtsSz}
    \centering
    \includegraphics[width=\textwidth]{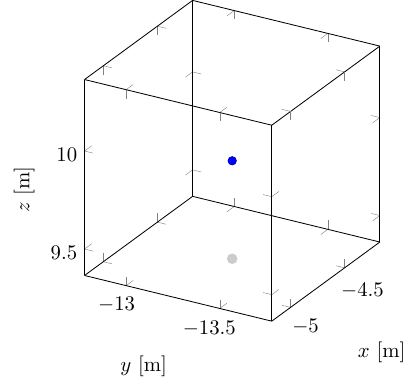}
    \caption{%
      UAV, $l = \SI{30}{\metre}$, 1 point.%
    }
  \end{subfigure}%
  ~
  \begin{subfigure}[t]{\detPtsSz}
    \centering
    \includegraphics[width=\textwidth]{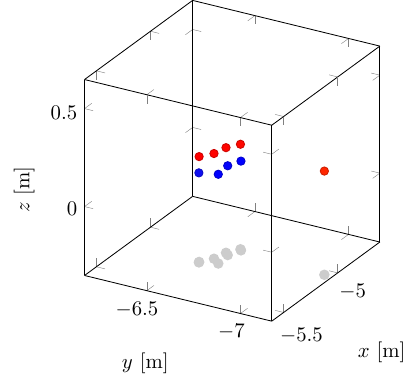}
    \caption{%
      Ground, $l = \SI{13}{\metre}$, 9 points.%
    }
  \end{subfigure}%

  \caption{%
    Example clusters of points corresponding to the MRS T650 UAV measured by the Ouster OS1-128 \ac{LiDAR} sensor at various distances.
    An example cluster of ground points is shown in the last image for comparison.
    The proposed detection algorithm can classify such point clusters even at longer distances with a low number of points and thus ambivalent spatial features.
  }
  \label{fig:detection_points}
\end{figure*}


Because the problem of low saliency of spatial features at long ranges in the task of \ac{UAV} detection from \ac{LiDAR} data is common to all state-of-the-art deep learning approaches, we conclude that contemporary neural networks are not suitable for this task.
For close objects, a deep neural network could be useful e.g. for detection confirmation or for resolving association ambiguities during the multi-target tracking.
However, this may be impractical when deployed onboard \ac{SWaP}-constrained \acp{UAV} due to the hardware and computational requirements.
}



\section{Real-world deployments}
To further highlight the performance of the system and its practicality, this section presents two cases of real-world deployment where the proposed detector and tracker were used to realize a complex, interacting multi-robot system.
Related videos are available online\footnote{\href{https://mrs.felk.cvut.cz/flying-object-detection}{\url{https://mrs.felk.cvut.cz/flying-object-detection}}}.
These use-cases are presented only briefly as a more in-depth description is beyond the scope of this paper.

\subsection{Deployment in cooperative navigation}
In \cite{pritzl2022icuas, pritzlFusionVisualInertialOdometry2023} and \cite{pritzl2023icra_droneguiding}, a cooperative navigation and exploration system for \acp{UAV} is presented.
A leader \ac{UAV} is equipped with a \ac{LiDAR} sensor and an onboard PC, which provide accurate navigation and occupancy mapping of the environment.
A light-weight follower \ac{UAV} equipped only with an onboard computer and a visual camera for navigation and obstacle avoidance is detected and tracked by the leader using the algorithms described in this paper.
Output of the tracking algorithm is aggregated over time and used to dynamically update the transformation between the follower's drifting self-localization and the leader's precise self-localization.
The leader periodically plans collision-free exploration trajectories for both \acp{UAV} based on the current detected position of the follower, which are transformed to their respective localization frames, and autonomously executed by the team-members.

\begin{figure}
  \centering
  \begin{subfigure}[t]{0.24\textwidth}
    \centering
    \includegraphics[width=\textwidth]{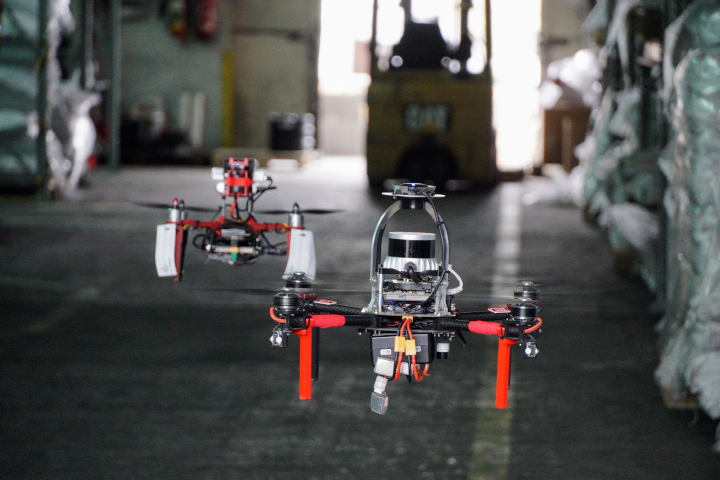}
    \caption{%
      Warehouse environment.
    }
  \end{subfigure}%
  ~%
  \begin{subfigure}[t]{0.24\textwidth}
    \centering
    \includegraphics[width=\textwidth]{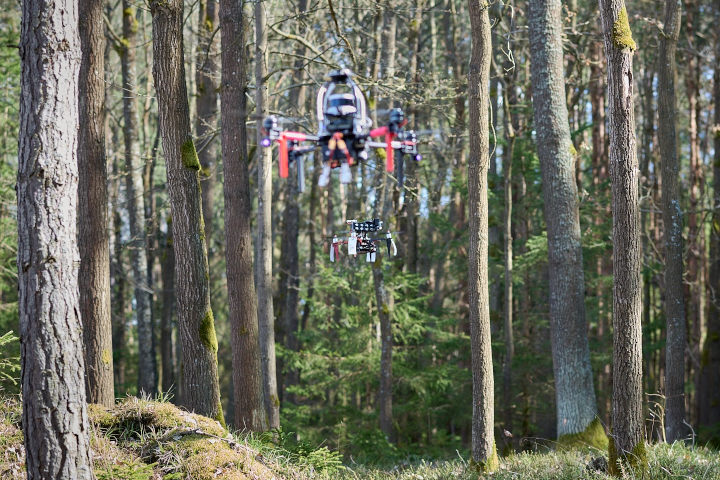}
    \caption{%
      Forest environment.
    }
  \end{subfigure}%
  \caption{
    Deployment of the proposed detection and tracking algorithms as part of a cooperative navigation multi-robot system in various environments~\cite{pritzl2023icra_droneguiding}.
  }
  \label{fig:coop_photos}
\end{figure}

A highly accurate position estimation of the detected target and the robust tracking algorithm that prevents false-positives and misassociations are crucial for the reliable functioning of such an approach.
The cooperative navigation system was successfully deployed in various scenarios both indoors and outdoors (see Fig.~\ref{fig:coop_photos}), which shows the robustness of the detection and tracking algorithms.


\subsection{Deployment in autonomous aerial interception}
\begin{figure}
  \centering
  \includegraphics[width=0.48\textwidth]{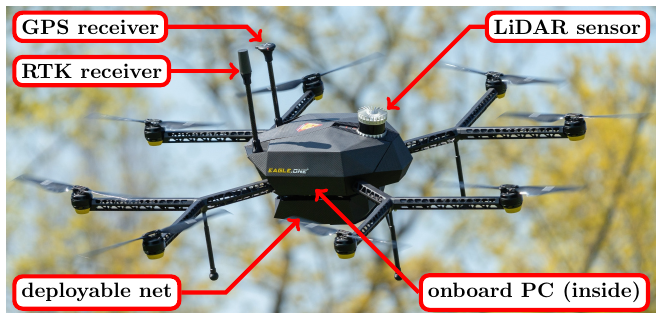}
  \caption{%
    A schematic of the autonomous aerial interception platform Eagle.One Mk4 during deployment.
  }
  \label{fig:eaglemk4_hw}
\end{figure}

The Eagle.One\footnote{\href{https://eagle.one}{\url{https://eagle.one}}, \href{https://mrs.felk.cvut.cz/projects/eagle-one}{\url{https://mrs.felk.cvut.cz/projects/eagle-one}}} is an \ac{AAIS} prototype being developed by our research group~\cite{vrba_ral2019, vrba_ral2020, pliska2024towards}.
During our research of autonomous aerial interception, we have identified the lack of a sufficiently fast, accurate, and robust detection system as the main obstacle for a practical realization of such a system, which was the main motivation for development of the detection and tracking methods presented in this paper.
These newly developed methods have been integrated into the Eagle.One Mk4 \ac{AAIS}, which has a similar hardware and software configuration as the \acp{UAV} used in sec.~\ref{sec:experiments} (see Fig.~\ref{fig:eaglemk4_hw}).

The system assumes a pre-scanned map of the environment for initialization of the detector and uses a slightly modified variant of the tracker with a more complex motion model based on an Interacting Multiple Model filter to accommodate the maneuvering of the target.
The target's state estimated by the filter is used by a specialized navigation algorithm based on Proportional Navigation.
Using these newly developed methods, we have successfully implemented an \ac{AAIS} capable of autonomously and reliably capturing a maneuvering target flying up to \SI{5}{\metre\per\second} (see videos\footnote{\href{https://mrs.felk.cvut.cz/flying-object-detection}{\url{https://mrs.felk.cvut.cz/flying-object-detection}}})~\cite{pliska2024towards}.
This further demonstrates the performance of the proposed detection and tracking algorithms, even under harsh conditions and when it is deployed in feedback with motion planning of the platform.




\section{Conclusion}
This paper has presented a new method for marker-less onboard detection and state estimation of flying objects.
The detection relies on occupancy mapping with a \ac{LiDAR} sensor using a novel algorithm that explicitly maps not only the occupied, but also free and unknown space, and takes into account dynamic objects.
A cluster of points measured by the sensor in an area mapped as confidently free is then recognized as a flying object by the detector.
A novel clustering multi-target tracking algorithm processes these detections, solves track association, and provides state estimation with low delay and computational complexity.
This makes the system suitable for deployment on board \acp{UAV} in both cooperative and non-cooperative multi-robot scenarios.

Theoretical limitations of the proposed method regarding the mapping accuracy and target position estimation are analyzed, providing the user with a powerful tool to estimate the expected performance given the limitations of a specific sensor or self-localization system.
The theoretical analysis is also applicable to general occupancy mapping or target position estimation using a sensor with uncertain pose, which, to the best of our knowledge, is an overlooked problem.
We believe the results will prove to be useful beyond the context of this paper.

Furthermore, a thorough evaluation both in simulated and real-world experiments demonstrates the robustness of the approach and good performance under realistic conditions.
This is further highlighted by the successful deployment of the detection and tracking system in two demanding use-cases: a multi-\ac{UAV} cooperative navigation, and autonomous aerial interception, which could not be easily performed using the current state-of-the-art methods.
The presented method enables a transfer of many robotic systems from controlled laboratory environments to practical deployment in challenging environments, which is why we release the source codes as publicly available for the robotic community\footnote{\href{https://mrs.felk.cvut.cz/flying-object-detection}{\url{https://mrs.felk.cvut.cz/flying-object-detection}}}.


\bibliographystyle{IEEEtran}
\bibliography{citations}



\clearpage

\begin{IEEEbiography}[{\includegraphics[width=1in,height=1.25in,clip,keepaspectratio]{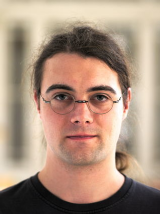}}]{Matouš Vrba}
received his M.Sc. degree in robotics and cybernetics at the Czech Technical University in Prague, Czech Republic, and is currently finishing his Ph.D. as a member of the \href{http://mrs.felk.cvut.cz/}{Multi-robot Systems lab} in CTU Prague since 2018.
His research focuses on machine perception methods for marker-less mutual localization of Unmanned Aerial Vehicles.
He is a co-author of 15 publications in conferences and impacted journals with $>\hspace{-0.5em}700$ citations indexed by Scholar and h-index 12.
He was also a member of CTU-UPENN-NYU team in the \href{http://mrs.felk.cvut.cz/mbzirc2020}{MBZIRC 2020}, and the CTU-CRAS-NORLAB team in the \href{http://mrs.felk.cvut.cz/projects/darpa}{DARPA SubT} competition.
\end{IEEEbiography}

\vspace{1em}

\begin{IEEEbiography}[{\includegraphics[width=1in,height=1.25in,clip,keepaspectratio]{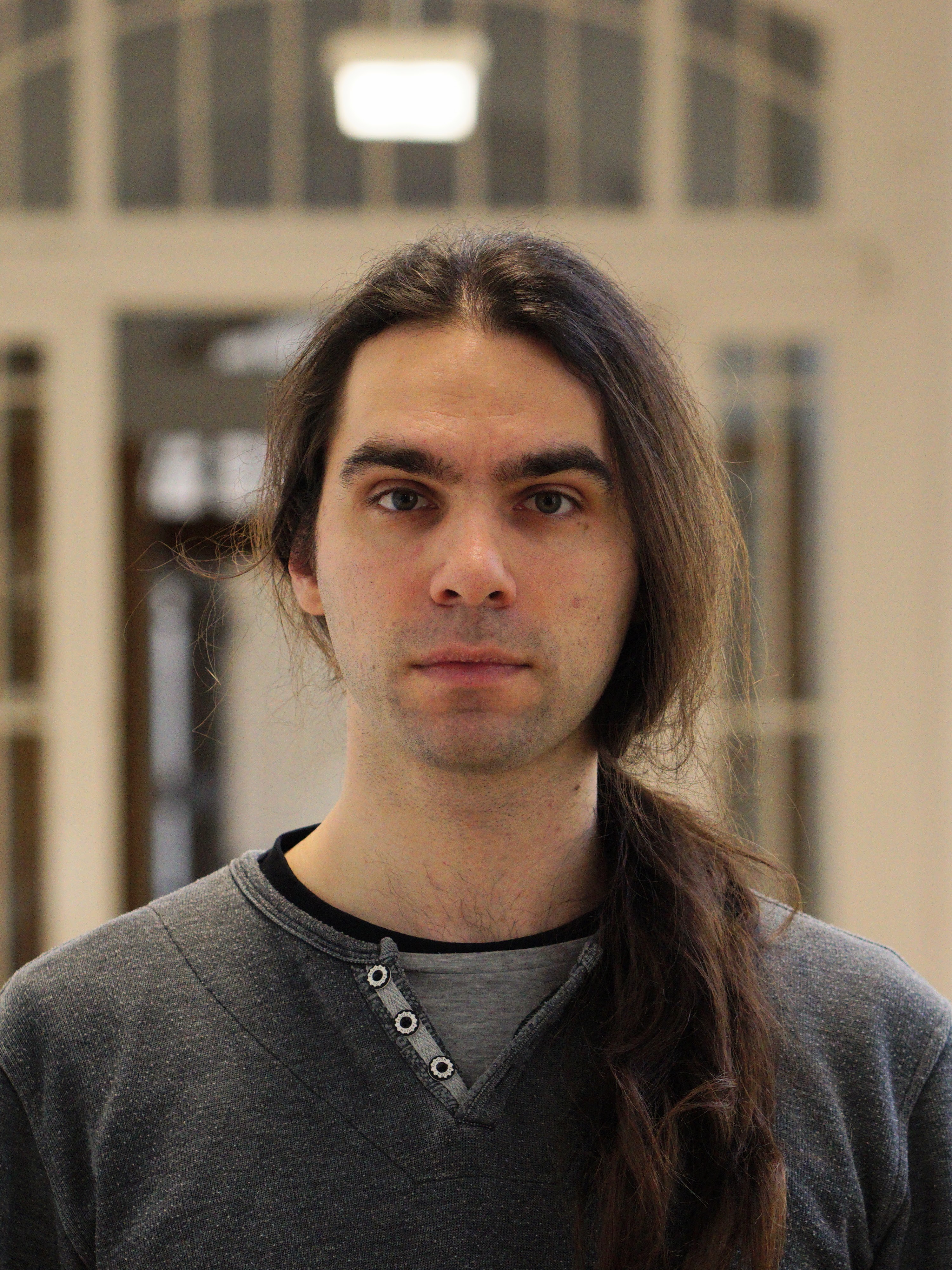}}]{Viktor Walter}
received his Ph.D. in robotics and cybernetics at the Czech Technical University in Prague, Czech Republic, and is currently employed in the \href{http://mrs.felk.cvut.cz/}{Multi-robot Systems lab} in CTU Prague.
The focus of his research is computer vision for mutual localization of Unmanned Aerial Vehicles, and swarming of aerial robots.
He is a co-author of 15 publications in conferences and impacted journals with $>\hspace{-0.4em}700$ citations indexed by Scholar and h-index 13.
He was also a member of CTU-UPENN-NYU team in the \href{http://mrs.felk.cvut.cz/mbzirc2020}{MBZIRC 2020}.
\end{IEEEbiography}

\vspace{0.5em}

\begin{IEEEbiography}[{\includegraphics[width=1in,height=1.25in,clip,keepaspectratio]{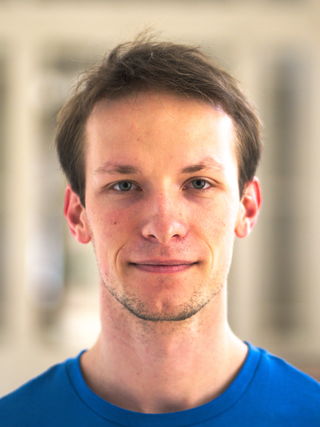}}]{Václav Pritzl}
received his M.Sc. degree in robotics and cybernetics at the Czech Technical University in Prague, Czech Republic, and is currently pursuing his Ph.D. as a member of the \href{http://mrs.felk.cvut.cz/}{Multi-robot Systems lab} in CTU Prague.
He is a co-author of 10 publications in conferences and impacted journals with $>\hspace{-0.4em}200$ citations indexed by Scholar and h-index 8.
The focus of his research is cooperative navigation of teams of Unmanned Aerial Vehicles in GNSS-denied environments, and he was also a member of CTU-UPENN-NYU team in the \href{http://mrs.felk.cvut.cz/mbzirc2020}{MBZIRC 2020}.
\end{IEEEbiography}

\vspace{1.5em}

\begin{IEEEbiography}[{\includegraphics[width=1in,height=1.25in,clip,keepaspectratio]{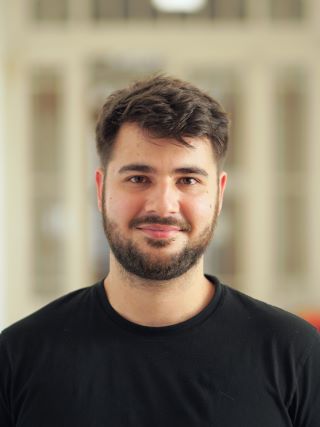}}]{Michal Pliska}
received the M.Sc. degree in cybernetics and robotics from Czech Technical University in Prague, Czechia, in 2023.
He is currently working toward the Ph.D. degree at the \href{http://mrs.felk.cvut.cz/}{Multi-robot Systems lab} in CTU Prague, supervised by Martin Saska.
He is the first author of 2 publications in impacted journals.
His research interests include multimodal state estimation and behavior prediction of UAVs, event-based vision, and planning through reinforcement learning.
\end{IEEEbiography}

\vfill

\newpage

\begin{IEEEbiography}[{\includegraphics[width=1in,height=1.25in,clip,keepaspectratio]{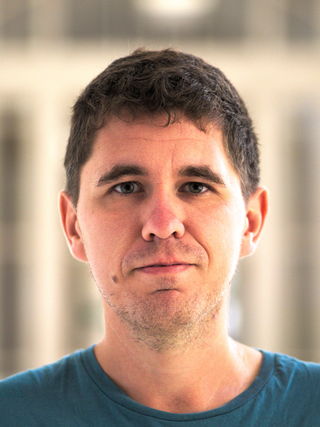}}]{Tomáš Báča}
received his Ph.D. degree on distributed remote sensing at the Czech Technical University in Prague, Czech Republic.
He is a part of the \href{http://mrs.felk.cvut.cz/}{Multi-robot Systems lab} in CTU Prague, focusing on distributed radiation sensing with Unmanned Aerial Vehicles and planning and control for Unmanned Aerial Vehicles.
Tomáš is a co-author of \mbox{$>\hspace{-0.5em}50$} publications in conferences and impacted journals with $>\hspace{-0.5em}3100$ citations indexed by Scholar and h-index 32.
He was a member of CTU-UPenn-UoL and CTU-UPENN-NYU teams in the \href{http://mrs.felk.cvut.cz/projects/mbzirc}{MBZIRC 2017} and \href{http://mrs.felk.cvut.cz/mbzirc2020}{MBZIRC 2020} robotic competitions in Abu Dhabi, and of the CTU-CRAS-NORLAB team in the \href{http://mrs.felk.cvut.cz/projects/darpa}{DARPA SubT} competition.
\end{IEEEbiography}

\begin{IEEEbiography}[{\includegraphics[width=1in,height=1.25in,clip,keepaspectratio]{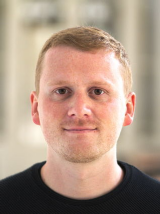}}]{Vojtěch Spurný}
received his Ph.D. degree on methods of planning and coordination for Unmanned Aerial Vehicles at the Czech Technical University in Prague, Czech Republic.
He is a part of the \href{http://mrs.felk.cvut.cz/}{Multi-robot Systems lab} in CTU Prague since 2014.
He is a co-author of \mbox{$>\hspace{-0.3em}20$} publications in conferences and impacted journals with $>\hspace{-0.3em}1600$ citations indexed by Scholar and h-index 18.
He was also a CTU-UPenn-UoL and CTU-UPENN-NYU team member in the \href{http://mrs.felk.cvut.cz/projects/mbzirc}{MBZIRC 2017} and \href{http://mrs.felk.cvut.cz/mbzirc2020}{MBZIRC 2020} robotic competitions in Abu Dhabi, United Arab Emirates.
\end{IEEEbiography}

\begin{IEEEbiography}[{\includegraphics[width=1in,height=1.25in,clip,keepaspectratio]{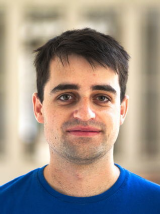}}]{Daniel Heřt}
is a robotics engineer specializing in unmanned aerial vehicles (UAVs) and multi-robot systems.
With a Master’s in robotics and cybernetics from the Czech Technical University in Prague, he leads hardware development for the \href{http://mrs.felk.cvut.cz/}{Multi-robot Systems lab}, focusing on UAV construction, electronics, and embedded software.
Dan has also contributed to nanosatellite projects at VZLU, and has published on UAV cooperative autonomy.
He was also a vital member of the CTU-UPenn-UoL and CTU-UPENN-NYU teams in the \href{http://mrs.felk.cvut.cz/projects/mbzirc}{MBZIRC 2017} and \href{http://mrs.felk.cvut.cz/mbzirc2020}{MBZIRC 2020} robotic competitions in Abu Dhabi, United Arab Emirates, and of the CTU-CRAS-NORLAB team in the \href{http://mrs.felk.cvut.cz/projects/darpa}{DARPA SubT} competition.
\end{IEEEbiography}

\begin{IEEEbiography}[{\includegraphics[width=1in,height=1.25in,clip,keepaspectratio]{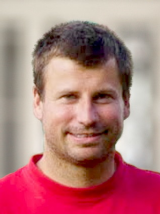}}]{Martin Saska}
received his Ph.D. degree at University of Wuerzburg, Germany, within the Ph.D. program of Elite Network of Bavaria.
He founded and heads the \href{http://mrs.felk.cvut.cz/}{Multi-robot Systems lab} at the Czech Technical University in Prague with more than 40 researchers.
He was a visiting scholar at University of Illinois at Urbana-Champaign and at University of Pennsylvania, USA.
He is a co-author of $>\hspace{-0.3em}200$ publications in conferences and impacted journals, including IJRR, AURO, JFR, ASC, EJC, with \mbox{$>\hspace{-0.3em}7700$} citations indexed by Scholar and h-index 48.
His team won multiple robotic challenges in \href{http://mrs.felk.cvut.cz/projects/mbzirc}{MBZIRC 2017}, \href{http://mrs.felk.cvut.cz/mbzirc2020}{MBZIRC 2020} and \href{http://mrs.felk.cvut.cz/projects/darpa}{DARPA SubT} competitions.
\end{IEEEbiography}

\vfill


\end{document}